\documentclass{article}

\usepackage{microtype}
\usepackage{graphicx}
\usepackage{booktabs} 

\usepackage{hyperref}

\usepackage[accepted]{icml2025}

\usepackage{subcaption}
\usepackage{graphicx}
\usepackage{caption}
\usepackage{amsmath}
\usepackage{amssymb}
\usepackage{mathtools}
\usepackage{amsthm}
\usepackage{colortbl}
\usepackage{multirow}
\usepackage{booktabs} 
\usepackage{arydshln}
\usepackage{amssymb}
\usepackage{pifont}
\usepackage{units}
\usepackage{xcolor} 
\usepackage{balance}
\usepackage{anyfontsize}
\usepackage{xurl} 

\usepackage{amsmath,amsfonts,bm}
\usepackage{mathtools}

\def\eqref#1{equation~\ref{#1}}

\def\1{\bm{1}}

\def\vq{{\bm{q}}}

\def\mI{{\bm{I}}}

\def\mK{{\bm{K}}}

\def\mP{{\bm{P}}}

\def\mR{{\bm{R}}}

\def\mU{{\bm{U}}}

\def\mW{{\bm{W}}}
\def\mX{{\bm{X}}}

\DeclareMathAlphabet{\mathsfit}{\encodingdefault}{\sfdefault}{m}{sl}
\SetMathAlphabet{\mathsfit}{bold}{\encodingdefault}{\sfdefault}{bx}{n}

\def\sR{{\mathbb{R}}}

\makeatletter
\renewcommand{\paragraph}{%
  \@startsection{paragraph}{4}%
  {\z@}{0.25ex \@plus 0.25ex \@minus .5ex}{-1em}%
  {\normalfont\normalsize\bfseries}%
}
\makeatother

\newcommand{\cmark}{\ding{51}}%
\newcommand{\xmark}{\ding{55}}%

\usepackage[capitalize,noabbrev]{cleveref}

\theoremstyle{plain}
\newtheorem{theorem}{Theorem}[section]

\newtheorem{lemma}[theorem]{Lemma}

\theoremstyle{definition}

\theoremstyle{remark}

\usepackage[textsize=tiny]{todonotes}

\icmltitlerunning{ResQ: Mixed-Precision Quantization of Large Language Models with Low-Rank Residuals}

\begin{document}

\twocolumn[
\icmltitle{ResQ: Mixed-Precision Quantization of Large Language Models\\ with Low-Rank Residuals}




\begin{icmlauthorlist}
\icmlauthor{Utkarsh Saxena}{purdue}
\icmlauthor{Sayeh Sharify}{dmx}
\icmlauthor{Kaushik Roy}{purdue}
\icmlauthor{Xin Wang}{dmx}
\end{icmlauthorlist}

\icmlaffiliation{purdue}{Department of Electrical and Computer Engineering, Purdue University, West Lafayette, USA}
\icmlaffiliation{dmx}{d-Matrix, Santa Clara, USA}

\icmlcorrespondingauthor{Utkarsh Saxena}{saxenau@purdue.edu}

\icmlkeywords{Machine Learning, ICML}

\vskip 0.3in
]



\printAffiliationsAndNotice{} 

\begin{abstract}
Post-training quantization (PTQ) of large language models (LLMs) holds the promise in reducing the prohibitive computational cost at inference time. Quantization of all weight, activation and key-value (KV) cache tensors to 4-bit without significantly degrading generalizability is challenging, due to the high quantization error caused by extreme outliers in activations. To tackle this problem, we propose \emph{ResQ}, a PTQ method that pushes further the state-of-the-art. By means of principal component analysis (PCA), it identifies a low-rank subspace (in practice $\nicefrac 1 8$ of the hidden dimension) in which activation variances are highest, and keep the coefficients within this subspace in high precision, e.g.~8-bit, while quantizing the rest to 4-bit. Within each subspace, invariant random rotation is applied to further suppress outliers.  We show that this is a provably optimal mixed precision quantization scheme that minimizes error. With the Llama and Qwen2.5 families of models, we demonstrate that ResQ outperforms recent uniform and mixed precision PTQ methods on a variety of benchmarks, achieving up to 33\% lower perplexity on Wikitext than the next best method \emph{SpinQuant}, and upto 3$\times$ speedup over 16-bit baseline. Code repository available 
\href{https://github.com/utkarsh-dmx/project-resq}{here}.
\footnote{\url{https://github.com/utkarsh-dmx/project-resq}}
\end{abstract}

\section{Introduction}
Growing capabilities of large language models (LLMs) come with an increasing computational cost at inference time. LLM inference has two distinct stages: \emph{prefilling}, which processes the input prompt and populates the internal state called KV (key-value) cache, and \emph{generation}, where tokens are generated autoregressively. The prefilling stage is compute-bound, requiring trillions of floating-point operations (FLOPs), whereas the generation stage is memory-bound due to iterative accesses and updates of the KV cache. These high computational costs are further amplified by modern LLMs' large sizes -- some exceeding 400 billion parameters -- and the increasingly long context lengths that necessitates large KV caches.
\begin{figure*}
    \includegraphics[width=\textwidth]{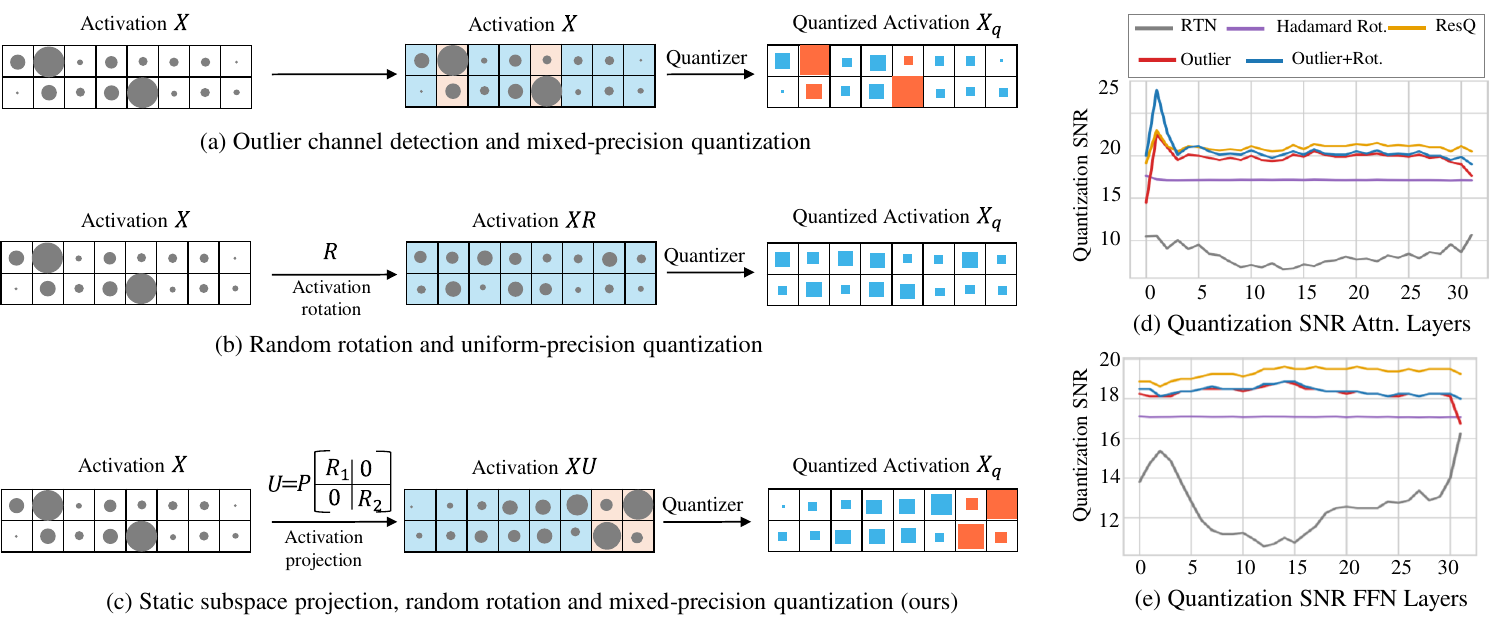}
    \caption{(a)-(c) Different approaches to quantization including ResQ. Symbol sizes represent magnitudes of values and colors indicate precisions of quantization (blue: low precision, orange: high precision). 
 (d)-(e) Quantization SNR comparison of ResQ with other baselines. }
    \label{fig1}
\end{figure*}

Quantization algorithms are powerful and principled approaches to address the immense computational demands of LLMs at both stages of inference. Quantization of weights reduces parameter storage, KV cache quantization mlowers memory usage of KV cache during generation, whereas activation quantization decreases the complexity of floating-point operation. However, effective low-precision quantization is difficult due to large outliers in activations, which can be $\sim20\times$ larger than other values \cite{dettmers2022gpt3}. While post-training methods like KIVI \cite{liu2024kivi} and KVQuant \cite{hooper2024kvquant} achieve 2-bit KV cache quantization, and techniques like GPTQ \cite{frantar2022gptq} and AWQ \cite{lin2024awq} optimize very low-precision weights, quantizing activations below 8-bit precision remains an open challenge. 

Recent LLM activation quantization methods feature two useful strategies: \emph{differential treatment of outliers} retain outlier channels in high precision, leading to mixed-precision quantization (e.g.,~\citealt{dettmers2022gpt3, zhao2024atom, ashkboos2023quik}; Figure \ref{fig1}a), whereas \emph{invariant random rotation} suppress outliers, leading to less difficult uniform low-precision quantization (e.g.,~\citealt{ashkboos2024quarot, liu2024spinquant}; Figure \ref{fig1}b).  Both reduce quantization error and improve signal-to-quantization-noise ratio (Figure \ref{fig1}d,e) locally; yet a notable model performance gap persists from the 16-bit baseline. For example, SpinQuant~\cite{liu2024spinquant} at 4-bit, applied to \texttt{Meta-Llama-3-8B} \cite{llama3meta}, exhibits $\sim20\%$ higher perplexity than the 16-bit floating point baseline, even after nontrivial optimization. 

To mend this gap, we introduce \emph{ResQ}, a novel PTQ method that combines the strengths of both aforementioned strategies and thereby improve model efficiency with aggressive 4-bit quantization of activation, weight, and KV cache. Specifically, by means of principal component analysis (PCA), we first identify a low-rank subspace that captures highest variances in activation, and mark the coefficients along this subspace for high-precision (8-bit) and the complement subspace for low-precision (4-bit) quantization. Then, ResQ employs invariant random rotations within each subspace before quantization to further suppress outliers~(Figure~\ref{fig1}c,d,e). We prove that the above treatment minimizes quantization error.  Similar to SpinQuant, most projection matrices can be fused into adjacent weights, leading to minimal runtime computational overhead (Section~\ref{sec:inference}). Furthermore, ResQ can be applied to KV cache quantization as well, and can be combined with GPTQ \cite{frantar2022gptq}, resulting in even better generalizing LLMs.

Outlier-based and rotation-based quantization methods can be combined. For example, high-precision outliers could be detected by $\ell_\infty$-norm similar to QUIK~\cite{ashkboos2023quik}, and random rotations applied within both high- and low-precision quantization groups, as in QuaRot~\cite{ashkboos2024quarot}. These methods fare less well than ResQ (Figure~\ref{fig1}d,e) in practice, in support of ResQ's provably optimal treatment of outlier quantization. When quantizing weight, activation and KV cache to 4-bit with only $\nicefrac 1 8$ channels in 8-bit, ResQ achieves 4-33\% lower perplexity on Wikitext and 0.1-5.4\% 0-shot accuracy improvements over SpinQuant~\cite{liu2024spinquant}, the best in practice so far. Unlike SpinQuant, ResQ does not require gradient-based optimization, making it a less demanding and faster PTQ solution. 
Furthermore, tuning the rank $r$ of ResQ gives rise to Pareto-optimal solutions as a tradeoff between efficiency and accuracy. 
We claim the following contributions.
\begin{enumerate}
\itemsep0em 
    \item We propose ResQ, a mixed precision weight, activation, and KV cache quantization method by keeping low-rank, high-variance components in high precision, in combination with random rotation-induced outlier suppression.
    \item We theoretically analyze the projection matrices in ResQ and show that using PCA-based projections minimizes quantization error. 
    \item We conduct extensive experiments on various models and language tasks and show that ResQ outperforms related state-of-the-art approaches.
    \item We develop CUDA kernels and achieve runtime speedup on NVIDIA GPUs with our quantized models.
\end{enumerate}
\section{Prior Work}\label{sec:priorwork}
\subsection{Quantization of LLMs}
Quantization reduces model size and accelerates inference by lowering neural network bit precision~\citep{choi2018pact, hubara2021accurate, yao2022zeroquant, park2022nuqmm, gholami2022survey, xi2023training}. It is broadly categorized into two categories: \emph{uniform precision quantization} (UPQ) and \emph{mixed precision quantization} (MPQ).
\textbf{Uniform precision quantization (UPQ)} applies the same bit-width across all layers, simplifying implementation but neglecting layer-specific sensitivity to quantization.  \textbf{Weight-only UPQ} methods reduce storage by compressing weights, using techniques like Hessian-guided rounding (GPTQ, \citealt{frantar2022gptq}), adaptive rounding (QuIP, \citealt{chee2024quip}), channel-wise scaling (AWQ, \citealt{lin2024awq}), and multi-codebook quantization (AQLM, \citealt{egiazarian2024extreme}). However, these methods struggle with batch processing due to significant activation memory overhead. \textbf{Weight-activation UPQ} compresses both weights and activations to address this. Methods such as SmoothQuant~\citep{xiao2023smoothquant} and OmniQuant~\citep{shao2023omniquant} scale activations and weights to handle outliers, while RPTQ~\citep{yuan2023rptq}, QLLM~\citep{liu2023qllm}, and QServe~\citep{lin2024qserve} employ channel-level strategies like clustering and reordering. Rotation-based methods such as QuaRot~\citep{ashkboos2024quarot}, SpinQuant~\citep{liu2024spinquant} and DuQuant \cite{lin2024duquant} further enhance robustness in low-precision scenarios. \textbf{KV cache UPQ} reduces memory for large batches or long contexts. FlexGen~\citep{sheng2023flexgen} employs 4-bit quantization and memory offloading, while KIVI~\citep{liu2024kivi} uses asymmetric 2-bit quantization for compression, enabling efficient inference.
\paragraph{Mixed precision quantization (MPQ)} optimizes bit-widths by adapting to the sensitivity of weights and activations, achieving better accuracy than UPQ at similar compression rates. \emph{Our proposed method, ResQ, follows the MPQ approach.} \textbf{Weight-only MPQ} has advanced efficiency for memory-bound applications with minimal activation demands. Methods like OWQ~\citep{lee2024owq} and SpQR~\citep{dettmers2023spqr} mitigate activation outliers' impact by retaining critical features in full precision, while SqueezeLLM~\citep{kim2023squeezellm} employs Dense-and-Sparse decomposition to efficiently store sensitive weights. \textbf{Weight-activation MPQ} enhances efficiency by addressing activation outliers (e.g.~\citealt{guan2024aptq, zeng2024abq}). Methods like LLM.int8()~\citep{dettmers2022gpt3} and QUIK\citep{ashkboos2023quik} preserve critical activations with mixed or low-precision decompositions, while Atom~\citep{zhao2024atom} and SliM-LLM~\citep{huang2024slim} optimize quantization through channel reordering and salience-driven bit allocation. 
\textbf{KV cache MPQ} reduces memory usage while preserving precision for critical tokens using techniques like non-uniform quantization, importance-aware precision, and salient token compression~\citep{hooper2024kvquant, yang2024notoken, dong2024qaq, he2024zipcache}. Alternatively, GEAR quantizes all tokens' KV cache and maintains low-rank quantization error~\citep{kang2024gear}.
\subsection{Low-rank decomposition}
Low-rank decomposition techniques have been widely used in model compression, reducing dimensionality while maintaining performance. For instance, SliceGPT~\cite{ashkboos2024slicegpt} projects weight matrices onto principal components for sparsification, while ESPACE~\cite{sakr2024espace} reduces activation dimensionality via pre-calibrated projections, achieving inference-time efficiency. Similarly, ASVD~\cite{yuan2023asvd} introduces an activation-aware decomposition method that incorporates activation distributions into weight decomposition.
Additionally, low-rank decomposition can be applied to reduce KV cache size. For example, Eigen Attention~\cite{saxena2024eigen} and ASVD~\cite{yuan2023asvd} employ low-rank approximations to reduce memory usage in KV caches during attention operations. PALU~\cite{chang2024palu} introduces learnable projections to adaptively compress KV caches based on the compression budget. Finally, Matryoshka KV Cache refines this with hierarchical orthogonal projections and knowledge distillation~\cite{lin2024matryoshkakv}.
\section{Quantization}
Quantization of weight, activation or KV cache involves converting component elements to low precision so that they can be represented using fewer bits for more efficient compute and storage. 
The $N$-bit integer quantization and dequantization process on matrix $\mX$ is given as 
\begin{equation}
    Q_N(\mX) = \Big\lfloor\frac{\mX - z_X}{s_X}\Big\rceil\cdot s_X + z_X ,
    \label{eq:quant}
\end{equation}
where $\lfloor\cdot\rceil$ is a round-and-clip function; $s_X$ and $z_X$ the scale and zero-point; $z_X = 0$, $s_X = \frac{\max(|\mX|)}{2^{N-1}-1}$ for symmetric quantization or $z_X = \min(\mX)$, $s_X = \frac{\max(\mX) - \min(\mX)}{2^N -1}$ for asymmetric quantization. 
\section{ResQ}
In this section, we introduce ResQ, a mixed-precision quantization approach that projects weights, activations, and the KV cache into an orthogonal space, retaining the low-rank components in high precision (8-bit) and the rest in low precision. We describe the quantization scheme, the generation of the basis space, provide theoretical guarantees, and outline end-to-end LLM inference deployment procedure.
\subsection{Quantization scheme}
\begin{figure}[t!]
    \includegraphics[width=\linewidth]{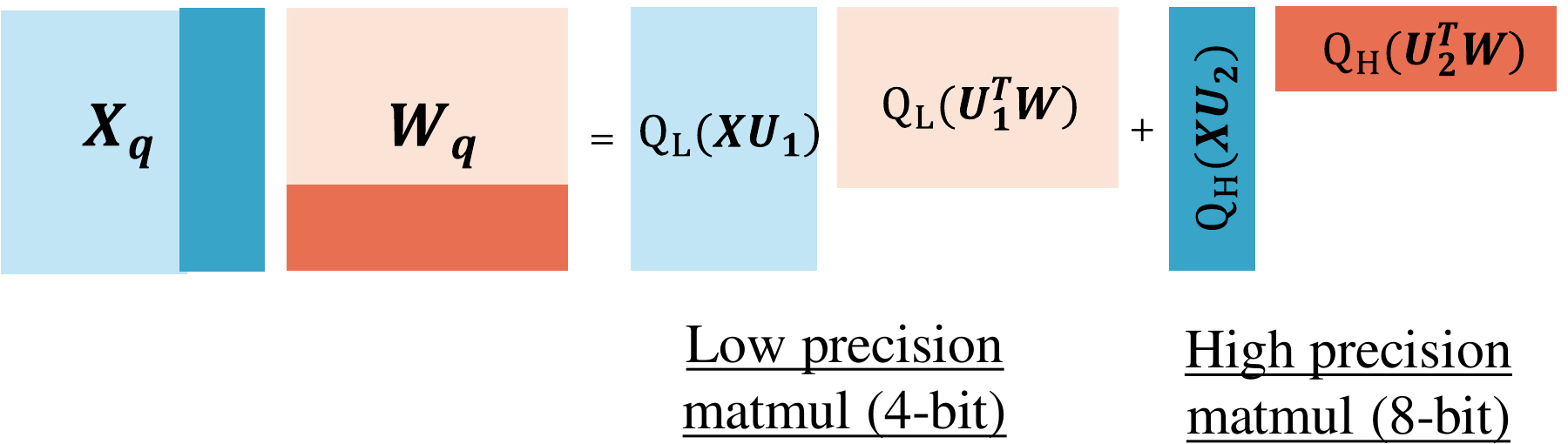}
    \caption{Matrix multiplication with mixed precision operands}
    \label{fig:matmul}
\end{figure}

Given input activation $\mX \in \sR^{n \times d}$ and weight $\mW \in \mathbb{R}^{d \times d}$, they are first projected onto an orthogonal basis defined by the vectors $\mU \in \sR^{d \times d}$. The coefficients of the projections along this basis are then subject to quantization. We seek to quantize some coefficients along certain bases at high precision while those remaining at low precision. Within $\mathbb{R}^{d}$, denote bases of a low-rank space of high-precision components by $\mU_{h} \in \sR^{d \times r}$ and those of its complementary subspace of low-precision components by $\mU_{l} \in \sR^{d \times (d-r)}$. The rank $r$ controls the amount of components in high precision (in practice we typically choose $r=\nicefrac{d}{8}$). We have $\mU_{h}\mU^\top_{h} + \mU_{l}\mU^\top_{l}  = \mU\mU^\top = \mI$ because $\mU$ is orthogonal. The quantized activation ${\mX_q}$ is thusly
\begin{equation}
    {\mX_q} =  Q({\mX\mU}) = Q_L({\mX\mU_{l}}) + Q_H({\mX\mU_{h}})
    \label{eq:act_quant}
\end{equation}
Similarly, quantized weights $\mW_q$ is obtained by projecting the inputs space of weights by $\mU^\top$ and quantizing the coefficients,
\begin{equation}
    {\mW_q} = Q({\mU^\top\mW}) = Q_L({\mU^\top_l\mW}) + Q_H({\mU^\top_h\mW}) .
\end{equation}
And the output of the layer is, 
\begin{equation}
    \begin{split}
    {\mX_q \mW_q} &= Q_L({\mX\mU_l})Q_L({\mU^\top_l\mW}) \\
    &+ Q_H({\mX\mU_h})Q_H({\mU^\top_h\mW}) .
    \end{split}
    \label{eq:layer_op}
\end{equation}
We make two observations due to orthogonality. First, \emph{the introduction of the projections do not alter the output of the model at infinite precision}. This means that, if quantization operation is removed from Equation~\ref{eq:layer_op}, the layer output is numerically invariant. Second, \emph{multiplication between low- and high-precision components vanishes (Figure~\ref{fig:matmul})}. This is efficient because only hardware kernels for quantized GEMM between operands of same precision are required.
\subsection{Projections and optimality thereof}
Intuitively, the orthogonal basis vectors $\mU$ should have two properties: (1) the low-rank space for high-precision quantization should capture the more important components, and (2) quantization error in both high- and low-precision groups should be minimized. We construct $\mU$ as a combination of two rotation matrices serving both objectives respectively. We write ${\mU_i = \mP_i\mR_i}, i \in {\{h,l\}}$. Therefore, 
\begin{equation}
    {\mU} = {\mP\mR} = [{\mP_l} \, {\mP_h}] \begin{bmatrix}
                        {\mR_l} & \mathbf{0} \\
                        \mathbf{0} & {\mR_h} 
                    \end{bmatrix} ,
\end{equation}
where, ${\mP_l,\mR_l}\in {\sR}^{d \times (d-r)}, {\mP_h,\mR_h}\in {\sR}^{d \times r}$. Inspired by prior work \cite{ashkboos2024quarot, chee2024quip}, we make ${\mR_l, \mR_h}$ random orthogonal matrices because random rotation reduces outliers, making the rotated matrices easier to quantize.  Furthermore, projection with a random orthogonal matrix increases Gaussianity of activations and weights within high- and low-precision groups, due to Lemma~\ref{lemma1}, conducive to the quantizations applied to these groups. 
\begin{figure}[t!]
    \includegraphics[width=\linewidth]{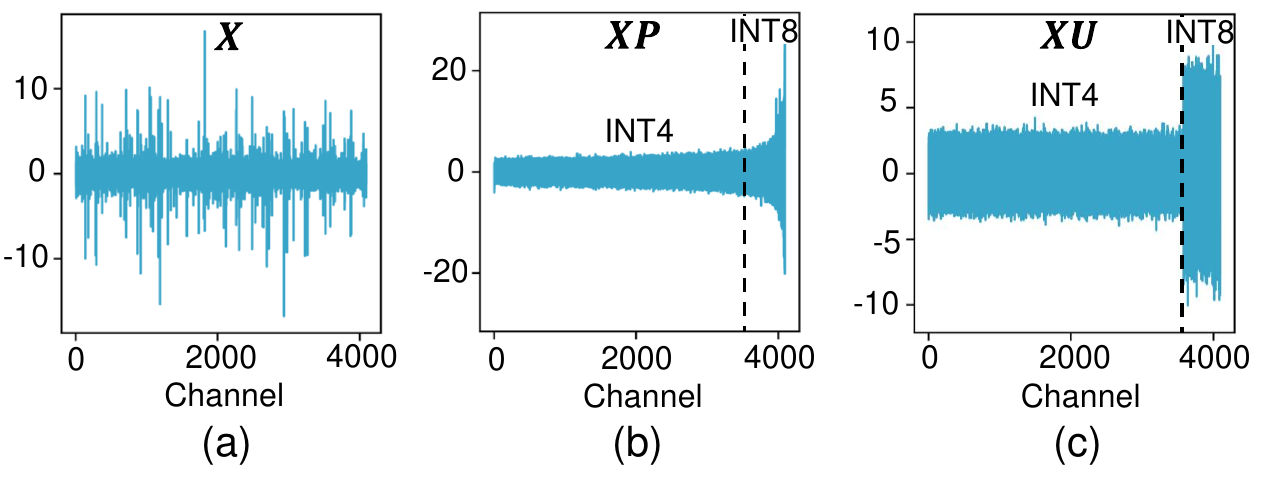}
    \caption{Activation distribution of the baseline and applying the projection matrices.}
    \label{fig:act_dist_small}
\end{figure}

\begin{figure*}[t!]
    \centering
    \includegraphics[width=\linewidth]{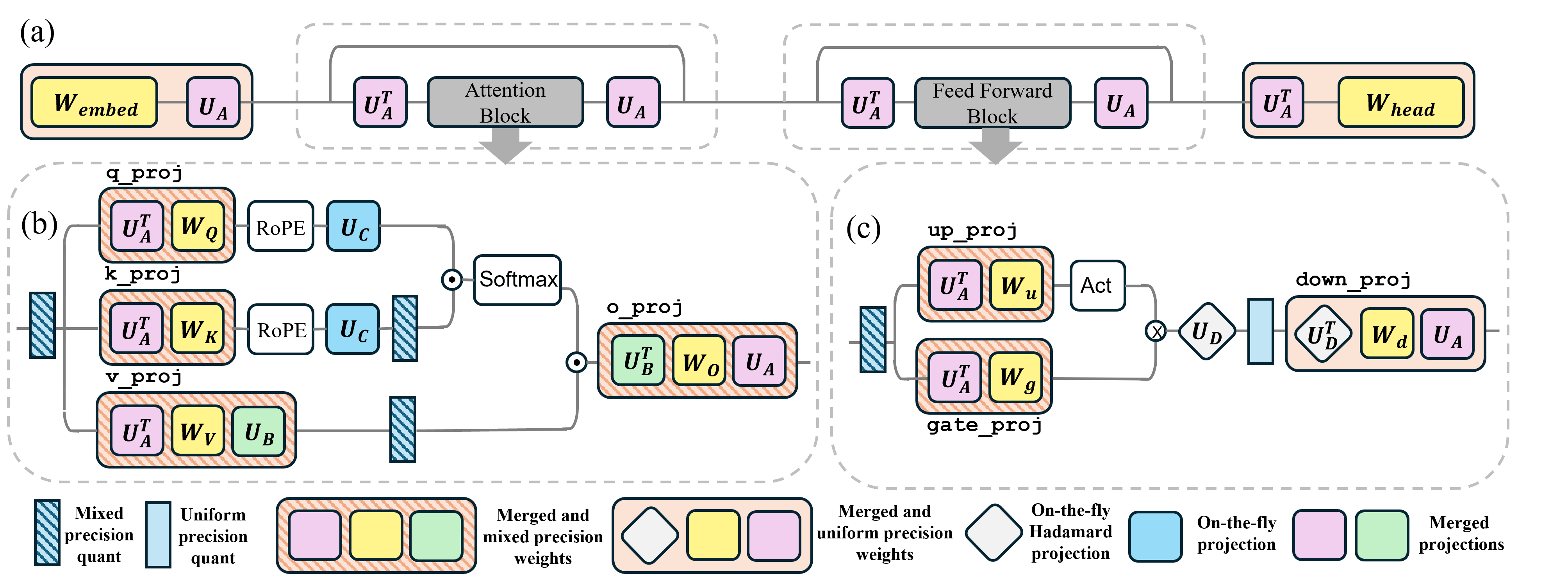}
    \caption{Model inference with ResQ incorporating the projection matrices. (a) $\mU_A$ modifies the inputs across blocks enabling better quantization. (b) $\mU_B, \mU_C$ enables mixed precision quantization of KV cache. (c) $\mU_D$ projects the activations and weights of \texttt{down\_proj} layer.}
    \label{fig:model_inference}
\end{figure*}
\begin{lemma}
By Central Limit Theorem, the distribution after multiplication with random orthogonal matrix is approximately Gaussian \cite{tseng2024quipsharp}.
\label{lemma1}
\end{lemma}
To determine ${\mP}$, we minimize the activation quantization error $\lVert{\mX} - {\mX_q}\rVert_F$.  
For activations quantized according to Equation~\ref{eq:act_quant}, we have,
\begin{equation}
    \begin{split}
        \lVert{\mX} - {\mX_q}\rVert_F &= \lVert{\mX\mU_l} - Q_L({\mX\mU_l})\rVert_F \\
        &+ \lVert{\mX\mU_h} - Q_H({\mX\mU_h})\rVert_F .
    \end{split}
    \label{eq:quant_error}
\end{equation}
\begin{theorem}
    For any matrix ${\mX}$ quantized to ${\mX_q}$ according to method described in Equation~\ref{eq:act_quant}, assuming the values to be quantized in ${\mX}$ are normally distributed, we have
    \begin{equation}
        \begin{split}
        {\mathbb{E}}\lVert\mX - \mX_q\rVert_F &\leq \frac{\sqrt{\pi{\log (d-r)}}}{2^{L-1}-1}\mathbb{E}\lVert{\mX}\rVert_F \\
        &-\left[\frac{\sqrt{\pi{\log (d-r)}}}{2^{L-1}-1} 
        - \frac{\sqrt{\pi{\log r}}}{2^{H-1}-1}\right]\\&\ \ \mathbb{E}\lVert{\mX\mP_h}\rVert_F .
        \end{split}
    \end{equation}
    \label{theorem1}
\end{theorem}
Full proof of Theorem~\ref{theorem1} is in Appendix~\ref{sec:proof_theorem}. Theorem~\ref{theorem1} bounds the quantization error in Equation~\ref{eq:quant_error} from above. To lower this upper bound of quantization error is thusly to maximize $\lVert\mX\mP_h\rVert_F$ which happens when $\mP_h$ comprises of eigenvectors of the covariance matrix $\mX\mX^\top$ with its \textit{largest} eigenvalues. Therefore, the low-rank subspace for high-precision quantization can be obtained by means of PCA, while the subspace for low-precision quantization can be obtained using ${\mU_h\mU^\top_h + \mU_l\mU^\top_l = \mP_h\mP^\top_h + \mP_l\mP^\top_l = \mI}$ (because ${\mR_i}$ is orthogonal). If we construct $\mP$ by taking eigenvectors of $\mX\mX^\top$ arranged in \emph{increasing} order of eigenvalues, the last $r$ columns of such a $\mP$ would correspond to $\mP_h$ and the first $d-r$ columns would correspond to $\mP_l$. The distribution of activation after applying different projection matrices is shown in Figure \ref{fig:act_dist_small}. Projection of activation along $\mP$ sorts the activation coefficients in increasing order of variance due to increasing eigenvalues of bases vectors. Consequently, the later $r$ channels of the projected activations with higher variance are kept in higher precision. Projection along $\mU = \mP\mR$ smoothes the activations along low precision and high precision groups further reducing quantization error (Figure \ref{fig:act_dist_small}) and improving quantization SNR (Figure \ref{fig1}(d,e)). 
\subsection{Inference computation with optimized projections}\label{sec:inference}
Once the projection matrices are obtained, the operation in Equation~\ref{eq:layer_op} requires multiplying the weights and activations with ${\mU}$. Weights can be projected and quantized offline. The projection operation on an activation can be merged to the weight of a previous linear layer. Based on the architecture of decoder based LLMs, we introduce four different kinds of projections (Figure~\ref{fig:model_inference}) : ${\mU_A \in \sR^{d_h \times d_h}}$, ${\mU_B, \mU_C \in \sR^{d_\text{head} \times d_\text{head}}}$, ${\mU_D \in \sR^{d_\text{FFN} \times d_\text{FFN}}}$ where $d_h$ is hidden dimension of LLM, $d_\text{head}$ is the attention head dimension and $d_\text{FFN}$ the hidden dimension of feedforward network (FFN).

\textbf{Projections at block boundaries}\ \ Input activations to attention and FFN are projected via $\mU_A$. Projection is handled by right-multiplying the weight matrix of final linear layer in each block (\texttt{o\_proj} in attention and \texttt{down\_proj} in FFN) by $\mU_A$. Thus, projections of activations is handled at no additional inference cost. To maintain numerical invariance, the first linear layer of each block (\texttt{q\_proj|k\_proj|v\_proj} in attention and \texttt{up\_proj|gate\_proj} in FFN) is pre multiplied with ${\mU_A^\top}$ (Figure~\ref{fig:model_inference}a). Similarly, the weights of the embedding layer and the final head are modified to manage projection of the residual stream.

\begin{table*}[t!]
\centering
\caption{Comparison of perplexity score on Wikitext, average 0-shot common sense reasoning accuracy and average 0-shot MMLU accuracy at \textbf{W/A/KV = 4/4/4} bits. Results of all techniques were obtained using their official codebase. Our work ResQ and QUIK \cite{ashkboos2023quik} keep $\nicefrac{1}{8}$ channels in 8-bit. All techniques except RTN employ GPTQ \cite{frantar2022gptq} for weight quantization. $\uparrow$ higher is better, $\downarrow$: lower is better. Full results in Appendix~\ref{sec:appendix_complete_results}, Tables~\ref{tab:llama2-7b_llama2-13b_eval_harness} and~\ref{tab:Qwen-A4W4KV4}.}
\label{tab:main_results}
\resizebox{0.9\textwidth}{!}{
\begin{tabular}{c|c|c|ccc|ccc}
\toprule 
 \multirow{2}{*}{\textbf{Family}}& \multirow{2}{*}{\textbf{Method}} & \textbf{Training}   & \multicolumn{3}{c}{\texttt{Meta-Llama-3-8B}} & \multicolumn{3}{c}{\texttt{Meta-Llama-3-70B}} \\ \cline{4-9}
 &  & \textbf{Free} & \textbf{Wiki ($\downarrow$)} & \textbf{Avg. 0-shot ($\uparrow$)} & \textbf{MMLU ($\uparrow$)} & \textbf{Wiki ($\downarrow$)} & \textbf{Avg. 0-shot ($\uparrow$)} & \textbf{MMLU ($\uparrow$)} \\ \hline \hline
 & 16-bit baseline     & \cmark & 6.1   & 67.1 & 63.1 & 2.9 & 73.1 & 75.9 \\ \cdashline{2-9}
 & RTN         & \cmark & 218.9 & 39.3 & 23.6 & 452.7 & 45.5 & 23.2 \\
 & GPTQ        & \cmark & 166.3 & 39.8 & 23.3 & 11.6e3 & 34.9 & 25.5 \\
 & SmoothQuant+ & \cmark & 78.2  & 42.5 & 24.7 & - & - & - \\
 & QUIK        & \cmark & 14.2  & 51.6 & 32.7 & 8.0 & 58.2 & 51.1 \\
 & QuaRot      & \cmark & 7.8   & 62.1 & 53.2 & 5.7 & 67.6 & 65.3  \\
 & SpinQuant   & \xmark & 7.4   & 63.8 & 56.2 & 6.2 & 65.7 & 59.4 \\
\multirow{-8}{*}{Llama 3} & \cellcolor[HTML]{CCFACC}ResQ & \cellcolor[HTML]{CCFACC}\cmark & \cellcolor[HTML]{CCFACC}\textbf{7.1} & \cellcolor[HTML]{CCFACC}\textbf{63.9} & \cellcolor[HTML]{CCFACC}\textbf{57.2} & \cellcolor[HTML]{CCFACC}\textbf{4.1} & \cellcolor[HTML]{CCFACC}\textbf{71.1} & \cellcolor[HTML]{CCFACC}\textbf{73.9} \\ \hline
 \multirow{2}{*}{\textbf{Family}}& \multirow{2}{*}{\textbf{Method}} & \textbf{Training}  & \multicolumn{3}{c}{\texttt{Llama-3.2-1B}} & \multicolumn{3}{c}{\texttt{Llama-3.2-3B}} \\ \cline{4-9}
 &  & \textbf{Free} & \textbf{Wiki ($\downarrow$)} & \textbf{Avg. 0-shot ($\uparrow$)} & \textbf{MMLU ($\uparrow$)} & \textbf{Wiki ($\downarrow$)} & \textbf{Avg. 0-shot ($\uparrow$)} & \textbf{MMLU ($\uparrow$)} \\ \hline \hline
 & 16-bit baseline     & \cmark & 9.8   & 54.9 & 36.9 & 7.8   & 62.7 & 54.8 \\ \cdashline{2-9}
 & RTN         & \cmark & 329.1 & 38.1 & 23.8 & 268.8 & 38.7 & 25.7 \\
 & GPTQ        & \cmark & 108.9 & 38.0 & 24.9 & 178.3 & 40.3 & 24.8  \\
 & SmoothQuant+ & \cmark & 228.9 & 38.0 & 24.1 & 96.1  & 39.0 & 25.9 \\
 & QUIK        & \cmark & 21.8  & 44.3 & 25.1 & 15.8  & 48.8 & 31.1 \\
 & QuaRot      & \cmark & 14.3  & 49.0 & 25.5 & 10.1  & 56.1 & 42.0 \\
 & SpinQuant   & \xmark & 13.6  & 48.8 & 25.6 & 9.2   & 57.9 & 44.2 \\
\multirow{-8}{*}{Llama 3.2} & \cellcolor[HTML]{CCFACC}ResQ & \cellcolor[HTML]{CCFACC}\cmark & \cellcolor[HTML]{CCFACC}\textbf{12.4} & \cellcolor[HTML]{CCFACC}\textbf{50.1} & \cellcolor[HTML]{CCFACC}\textbf{29.4} & \cellcolor[HTML]{CCFACC}\textbf{8.8} & \cellcolor[HTML]{CCFACC}\textbf{59.0} & \cellcolor[HTML]{CCFACC}\textbf{49.8} \\ \hline
\multirow{2}{*}{\textbf{Family}}&  \multirow{2}{*}{\textbf{Method}} & \textbf{Training}  & \multicolumn{3}{c}{\texttt{Qwen2.5-3B}} & \multicolumn{3}{c}{\texttt{Qwen2.5-72B}} \\ \cline{4-9}
 &  & \textbf{Free} & \textbf{Wiki ($\downarrow$)} & \textbf{Avg. 0-shot ($\uparrow$)} & \textbf{MMLU ($\uparrow$)} & \textbf{Wiki ($\downarrow$)} & \textbf{Avg. 0-shot ($\uparrow$)} & \textbf{MMLU ($\uparrow$)} \\ \hline \hline
 & 16-bit baseline      & \cmark & 8.0     & 63.8 & 66.1 & 3.9     & 73.4 & 84.3 \\ \cdashline{2-9}
 & RTN          & \cmark & 39033.0 & 35.1 & 23.4 & 45412.7 & 34.3 & 24.0 \\
 & GPTQ         & \cmark & 9977.8  & 35.1 & 23.2 & 37967.2 & 34.5 & 23.3 \\
 & SmoothQuant+ & \cmark & 73306.7 & 34.8 & 23.9 & - & - & - \\
 & QUIK         & \cmark & 15.5    & 51.2 & 39.4	& 8.3 & 61.9 & 69.3 \\
 & QuaRot       & \cmark & 68.8    & 47.7 & 28.9 & 4.9 & 70.3 & 80.1\\
\multirow{-7}{*}{Qwen2.5} & \cellcolor[HTML]{CCFACC}ResQ & \cellcolor[HTML]{CCFACC}\cmark & \cellcolor[HTML]{CCFACC}\textbf{9.0} & \cellcolor[HTML]{CCFACC}\textbf{61.1} & \cellcolor[HTML]{CCFACC}\textbf{61.2} & \cellcolor[HTML]{CCFACC}\textbf{4.6} & \cellcolor[HTML]{CCFACC}\textbf{72.0} & \cellcolor[HTML]{CCFACC}\textbf{81.5} \\ 

\bottomrule 
\end{tabular}
}
\end{table*}

\textbf{Projections within the attention block}\ \ ${\mU_B, \mU_C}$ ensures that activations within attention block are projected (Figure~\ref{fig:model_inference}b). Post-multiplication of value projection layer by ${\mU_B}$ ensures that value vectors in KV cache are projected and quantized optimally. Consequently, the weights of \texttt{o\_proj} layer need to be pre multiplied by ${\mU^\top_B}$ to ensure numerical invariance. $\mU_C$ ensures that the quantization of key in KV cache is handled optimally. To achieve that, it is required to project both the query and key using the same projection matrix ${\mU_C}$. The attention dot product  remains invariant under projected inputs, 
\begin{equation}
{\vq_\text{proj}\mK^\top_\text{proj} = (\vq\mU_C)(\mU_C^\top\mK^\top) = \vq\mK^\top},
\end{equation}
where $\vq$ and $\mK$ are query and key after rotary embedding (RoPE), respectively. Because ${\mU_C}$ cannot be merged into the previous linear layer due to presence of RoPE, the projection is explicitly computed at runtime, but made more efficient by applying uniform precision quantization to $\mU_C$ and corresponding input activations.  

\textbf{Projections within the feedforward block}\ \ ${\mU_D}$ ensure improved quantization of activation within FFNs (Figure~\ref{fig:model_inference}c). $\mU^\top_D$ is left-multiplied with weights of \texttt{down\_proj}, but due to the presence of activation functions within the block, ${\mU_D}$ cannot be merged to weights of preceding linear layers and is computed at runtime. $\mU_D$ is applied to the hidden dimension of the FFNs ($d_\text{FFN}$) which is typically 3$\times$ to 4$\times$ the embedding dimension in most LLMs. In this scenario, matrix multiplication with $\mU_D$ is extremely expensive in computation and storage. To minimize the overhead, we choose $\mU_D$ to be a hadamard matrix to leverage fast and efficient hadamard transform kernel. And, we choose weights and activations for \texttt{down\_proj} layer to be uniformly quantized to low precision.

\begin{table*}[t!]
\centering
\caption{Comparison of performance of quantization approaches on generative tasks at precisions of \textbf{W/A/KV = 4/4/4} bits. Our work ResQ and QUIK \cite{ashkboos2023quik} keep $\nicefrac{1}{8}$ of channels in 8-bit.}
\label{tab:generative_benchmarks}
\resizebox{0.8\textwidth}{!}{
\begin{tabular}{c|c|c|cc|ccc}
\hline
 &  &  & \multicolumn{2}{c|}{\textbf{GSM8K 5-shot ($\uparrow$)}} & \multicolumn{3}{c}{\textbf{LongBench ($\uparrow$)}} \\ \cline{4-8}
\multirow{-2}{*}{\textbf{Model}} & \multirow{-2}{*}{\textbf{Method}} & \multirow{-2}{*}{\begin{tabular}[c]{@{}c@{}}\textbf{Training} \\ \textbf{Free}\end{tabular}} & flexible extract & strict match & qmsum & samsum & repobench-p \\ \hline \hline
 & 16-bit    & \cmark & 51.0 & 50.6 & 23.9 & 44.8 & 66.4 \\ \cdashline{2-8}
 & QUIK      & \cmark & 2.3  & 0.0  & 10.5 & 25.2 & 37.6 \\
 & QuaRot    & \cmark & 27.6 & 27.1 & 22.0 & 43.8 & 60.6 \\
 & SpinQuant & \xmark & 29.8 & 29.6 & 23.0 & 43.9 & \textbf{62.6} \\
\multirow{-5}{*}{\texttt{Meta-Llama-3-8B}} & \cellcolor[HTML]{CCFACC}ResQ  & \cellcolor[HTML]{CCFACC} \cmark& \cellcolor[HTML]{CCFACC}\textbf{33.6} & \cellcolor[HTML]{CCFACC}\textbf{33.2} & \cellcolor[HTML]{CCFACC}\textbf{23.1} & \cellcolor[HTML]{CCFACC}\textbf{44.1} & \cellcolor[HTML]{CCFACC}62.3 \\ \hline
 & 16-bit    & \cmark & 25.1 & 24.9 & 23.1 & 43.0 & 64.4 \\ \cdashline{2-8}
 & QUIK      & \cmark & 2.5  & 0.0  & 15.9 & 31.7 & 30.9 \\
 & QuaRot    & \cmark & 10.1 & 9.1 & 20.6 & 39.5 & 56.8 \\
 & SpinQuant & \xmark & 11.6 & 11.4 &   \textbf{21.7}   &  41.9    &   59.1   \\
\multirow{-5}{*}{\texttt{Llama-3.2-3B}} & \cellcolor[HTML]{CCFACC}ResQ & \cellcolor[HTML]{CCFACC} \cmark& \cellcolor[HTML]{CCFACC}\textbf{17.1} & \cellcolor[HTML]{CCFACC}\textbf{16.7} & \cellcolor[HTML]{CCFACC}\textbf{21.7} & \cellcolor[HTML]{CCFACC}\textbf{43.0} & \cellcolor[HTML]{CCFACC}\textbf{61.5} \\ \hline
\end{tabular}
}
\end{table*}
\section{Experiments}
\subsection{Setup} \label{sec:experiments,setup}
\textbf{Models, tasks, datasets and baselines}\ \ We conduct experiments on Llama 2~\cite{touvron2023llama2openfoundation}, Llama 3~\cite{llama3meta}, and the recently released Llama 3.2~\cite{llama3.2meta} and Qwen2.5~\cite{yang2024qwen2} models. We also include multi-modal language models belonging to Qwen2 VL family \cite{wang2024qwen2vl} for our evaluations. We benchmark our approach against GPTQ \cite{frantar2022gptq}, QuaRot \cite{ashkboos2024quarot}, QUIK \cite{ashkboos2023quik}, SpinQuant \cite{liu2024spinquant} and SmoothQuant+, a stronger baseline created by combining SmoothQuant \cite{xiao2023smoothquant} with GPTQ following \citealt{sharify2024ptq}. We evaluate the quantization approaches on a range of tasks which measure the 
\emph{language modeling ability}: perplexity on Wikitext~\cite{wikitext-arxiv2016}, \emph{common sense reasoning ability}: average 0-shot accuracy on Arc-c/e~\cite{arc-arxiv2018}, BoolQ~\cite{clark2019boolq}, HellaSwag~\cite{hellaswag-arxiv2019}, Openbook QA~\cite{OpenBookQA2018}, PIQA~\cite{piqa-aaai2020}, SIQA~\cite{sap2019social}, WinoGrande~\cite{winogrande-acm2021},  \emph{language understanding}: 0-shot accuracy on MMLU~\cite{hendryckstest2021mmlu},  \emph{mathematical understanding}: 5-shot GSM8K~\cite{cobbe2021traininggsm8k},  \emph{dialogue summarization}: samsum~\cite{gliwa2019samsum} and qmsum~\cite{zhong2021qmsum} from LongBench~\cite{bai2024longbench}, \emph{code completion}: repobench-p~\cite{liu2023repobenchbenchmarkingrepositorylevelcode} from LongBench, and \emph{multi-modal understanding}: MMMU \cite{yue2024mmmu}. 

\textbf{Implementation details}\ \ We implement ResQ using the HuggingFace Transformers library \cite{huggingface-arxiv2019} with PyTorch \cite{pytorch-Neurips2019}.  We share a single ${\mU_A}$ across all layers, while ${\mU_B}$, ${\mU_C}$ and ${\mU_D}$ are generated per layer. Following SpinQuant \cite{liu2024spinquant}, we use per-token asymmetric quantization for activations, per-channel symmetric quantization for weights, and per-head asymmetric quantization for the KV cache. We fuse the projection matrices ${\mU_A}, {\mU_B}, {\mU_D} $ into weights and apply GPTQ \cite{frantar2022gptq} for weight quantization. To efficiently implement on-the-fly projections, ${\mU_D}$ is a Hadamard matrix and $\mU_C$ and its activations are quantized to 8-bit. The entire process, including obtaining projections and quantization, runs on a single NVIDIA A100 GPU; for \texttt{Meta-Llama-3-8B}, it takes 35 minutes. Additional details are in Appendix \ref{sec:implementation_details}. 
\begin{table}[]
\centering
\caption{MMMU accuracy (higher is better) of vision language models when quantized using various approaches. For 4-bit data structures, our work ResQ and QUIK \cite{ashkboos2023quik} keep $\nicefrac{1}{8}$ of channels in 8-bit.}
\label{tab:vllm}
\resizebox{0.9\columnwidth}{!}{
\begin{tabular}{c|c|c|c}
\hline
 &  & \multicolumn{2}{c}{\textbf{Model}} \\ \cline{3-4} 
\multirow{-2}{*}{\textbf{W/A/KV} {(bit)}} & \multirow{-2}{*}{\textbf{Method}} & \multicolumn{1}{c|}{{\color[HTML]{242424} \texttt{\begin{tabular}[c]{@{}c@{}}Qwen2-VL\\ -2B-Instruct\end{tabular}}}} & {\color[HTML]{242424} \texttt{\begin{tabular}[c]{@{}c@{}}Qwen2-VL\\ -7B-Instruct\end{tabular}}} \\ \hline \hline
  {16/16/16} & Baseline & 39.6 & 51.6 \\ \cdashline{1-4}
             & RTN      & 25.0 & 26.7 \\
             & GPTQ     & 27.7 & 24.9 \\
             & QUIK     & 26.3 & 28.9 \\
             & QuaRot   & 24.0 & 24.5 \\
\multirow{-5}{*}{4/4/4} & \cellcolor[HTML]{CCFACC}ResQ & \multicolumn{1}{c|}{\cellcolor[HTML]{CCFACC}\textbf{29.7}} & \cellcolor[HTML]{CCFACC}\textbf{47.0} \\ \hline
 & RTN & 24.9 & 25.2 \\
 & GPTQ & 23.4 & 24.3 \\
 & QUIK & 28.4 & 26.4 \\
 & QuaRot & 26.5 & 24.5 \\
\multirow{-5}{*}{4/8/4} & \cellcolor[HTML]{CCFACC}ResQ & \multicolumn{1}{c|}{\cellcolor[HTML]{CCFACC}\textbf{34.0}} & \cellcolor[HTML]{CCFACC}\textbf{48.8} \\ \hline
\end{tabular}
}
\end{table}
\begin{table}[t]
\centering
\caption{Wikitext perplexity comparison of ResQ and baseline which keeps channels with high $l_\infty$-norm in high precision and uses rotation to reduce quantization error within high precision and low precision groups. }
\label{tab:outlier+rot}
\resizebox{0.9\columnwidth}{!}{
\begin{tabular}{c|c|c|c}
\hline
\textbf{W/A/KV} (bit) & \textbf{Method} & \texttt{\begin{tabular}[c]{@{}c@{}}Meta-Llama\\ -3-8B\end{tabular}} & \texttt{\begin{tabular}[c]{@{}c@{}}Llama\\ -3.2-3B\end{tabular}} \\
\hline \hline
\multirow{2}{*}{4/4/4} & outlier+rot & 7.2 & 9.0 \\
 & ResQ & \textbf{7.1} & \textbf{8.8} \\ \hline
\multirow{2}{*}{4/8/4} & outlier+rot & 6.6 & 8.3 \\
 & ResQ & \textbf{6.5} & \textbf{8.2} \\ \hline
\multirow{2}{*}{3/8/3} & outlier+rot & 7.7 & 9.9 \\
 & ResQ & \textbf{7.5} & \textbf{9.8} \\ \hline
\multirow{2}{*}{2/8/8} & outlier+rot & 12.6 & 16.0 \\
 & ResQ & \textbf{12.1} & \textbf{15.7} \\ \hline
\multirow{2}{*}{3/3/3} & outlier+rot & 14.7 & 18.7 \\
 & ResQ & \textbf{14.6} & \textbf{17.5} \\ \hline
\end{tabular}
}
\end{table}
\subsection{Main Results}
\textbf{Language modeling, understanding, and reasoning tasks}\ \ We evaluate ResQ on tasks that test language modelling ability (perplexity on Wikitext), common sense reasoning ability (average 0-shot accuracy on the eight tasks listed in section \ref{sec:experiments,setup}) and language understanding (average 0-shot accuracy on MMLU). The results are presented in Table \ref{tab:main_results}. We see that ResQ reduces the gap to 16-bit performance and outperforms the quantization baselines across all tasks on all models. Particularly, on Llama 3/3.2 family of models, ResQ outperforms SpinQuant by achieving 4-33\% lower Wikitext perplexity, 0.1-5.4\% better average 0-shot accuracy and a 1-14.5\% better accuracy on MMLU benchmark without any additional training. For the Qwen-2.5 model family, all other baselines fail to achieve competitive results, and ResQ significantly outperforms them. Compared with QUIK, another mixed precision quantization approach, ResQ achieves 42-50\%  better Wikitext perplexity, 5.8-12.3\% better average zero shot accuracy and 4.3-24.5\% better MMLU accuracy over all models. Complete results on Llama and Qwen2.5 family of models are provided in Appendix \ref{sec:appendix_complete_results}. Additionally, we provide comparison between baselines at \textbf{W/A/KV = 4/8/4} bits for Llama families in Appendix \ref{sec:appendix_w4a8kv4}. Among all the set of results, ResQ maintains superior performance. 

\textbf{Generative tasks}\ \ We also test ResQ on tasks that require auto-regressive token generation including the GSM8K mathematical understanding benchmark, dialogue summarization benchmarks (qmsum and samsum) and code completion benchmark (repobench-p, Table \ref{tab:generative_benchmarks}). The goal of choosing these tasks is to evaluate the generation ability on a wide variety of domains. On the challenging GSM8K benchmark where QUIK fails to produce meaningful results, ResQ outperforms SpinQuant by 3.8\% and 5.5\% on the 8B and 3B parameter model respectively, closing the gap to the 16-bit baseline. On LongBench evaluation tasks, ResQ demonstrates competitive performance and outperforms SpinQuant without any additional training. 

\textbf{Multi-modal understanding}\ \ We benchmark the quantization approaches on vision language models (VLMs) by quantizing Qwen2 VL family and evaluating their performance on MMMU (Table~\ref{tab:vllm}, \citealt{yue2024mmmu}). Only the language model is quantized while the vision encoder remains in 16-bit as the language model has many more parameters (over $10\times$ for \texttt{Qwen2-VL-7B-Instruct}). ResQ outperforms baselines on both 2B and 7B models, achieving superior accuracy and demonstrating its generalizability. Results for individual MMMU tasks are provided in Appendix~\ref{sec:appendix_MMMU}.

\textbf{Comparison against outliers with rotation baseline}\ \ A stronger baseline can be created combining existing quantization approaches. Like QUIK, one can find channels which consistently contain outliers and keep them in 8-bit while keep the remaining channels in low precision. And, the quantization of high/low precision groups can be improved using random rotations introduced in QuaRot. Compared with such a baseline which keeps channels with high $l_\infty$-norm in 8-bit, ResQ's unique approach involves keeping coefficients along bases with high eigenvalues in 8-bit. We see in Table \ref{tab:outlier+rot} that ResQ consistently outperforms such a strong baseline across various precisions of \textbf{W/A/KV} highlighting ResQ's PCA driven theoretically optimal approach of choosing high precision components.  
\begin{table}[t]
\centering
\caption{Impact of different projections in ResQ. Evaluated by removing components and observing Wikitext perplexity.}
\label{tab:ablation_projections}
\resizebox{\columnwidth}{!}{
\begin{tabular}{l|c|c|c|c|c|c}
\hline
             & \multirow{3}{*}{ResQ} &  \multicolumn{5}{c}{Removed Projections} \\ \cdashline{3-7}
             &                       & \multirow{2}{*}{${\mU_D}$} & \multirow{2}{*}{${\mU_A}$} & \multirow{2}{*}{${\mU_B}$} & \multirow{2}{*}{${\mU_C}$} & ${\mU_C}$, \\
             &              &       &      &     &     & ${\mU_B}$ \\ \hline \hline
\texttt{Llama-2-7b-hf}   & \textbf{5.8} & 1550  & 2500 & 5.8 & 5.9 & 5.9   \\
\texttt{Meta-Llama-3-8B}   & \textbf{7.1} & 1607  & 37.4 & 7.2 & 7.3 & 7.4   \\
\texttt{Llama-3.2-3B} & \textbf{8.8} & 279.2 & 39.0 & 9.0 & 9.2 & 9.4   \\ \hline
\end{tabular}
}
\end{table}
\subsection{Hardware performance}
\begin{figure}[t]
  \includegraphics[width=\columnwidth]{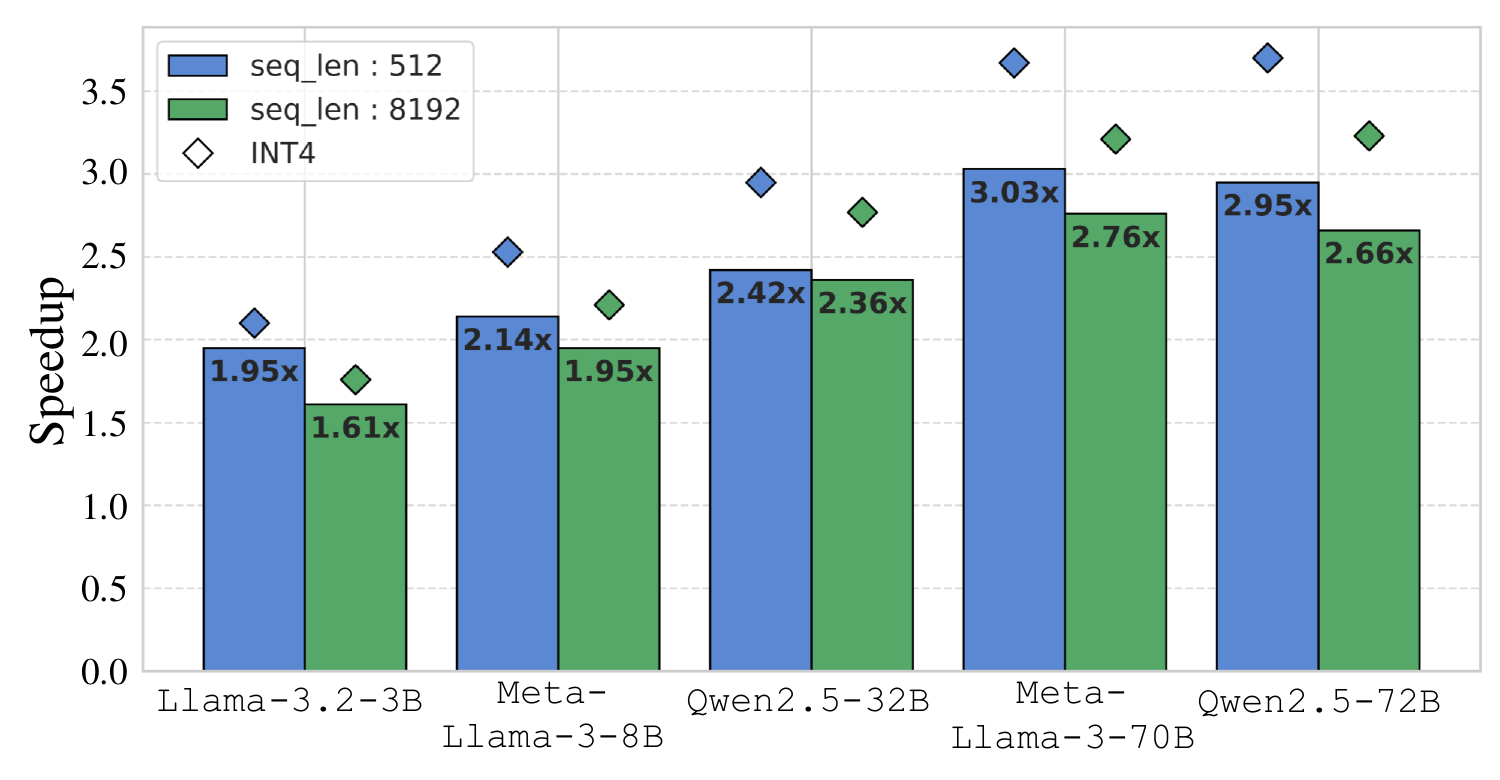}
  \caption{Speedup of ResQ and INT4 kernel on single decoder block on NVIDIA RTX 3090 over 16-bit floating point baseline for batch size of 1.}
  \label{fig:hardware}
\end{figure}

We implement the mixed-precision quantization using CUDA 11.8 and PyTorch. We use \texttt{CUTLASS} \cite{Thakkar_CUTLASS_2023} to perform \texttt{INT4} and \texttt{INT8} GEMM operations on TensorCore. On NVIDIA RTX 3090 GPU, representative of resource constrained setting, we achieve about $1.61\times$ to $3.03\times$ speedup with ResQ over 16-bit baseline for a single decoder block tested across several language models~(Figure~\ref{fig:hardware}). We observe higher speedup on larger models and smaller sequence lengths. Compared with INT4 implementation, ResQ kernel is only $14\%$ slower on average across models and sequence lengths demonstrating minimal overhead associated with mixed precision computations and on-the-fly projections introduced in ResQ. 
\begin{figure}[!t]
	\centering
	\begin{subfigure}{0.235\textwidth}
		\includegraphics[width=\textwidth]{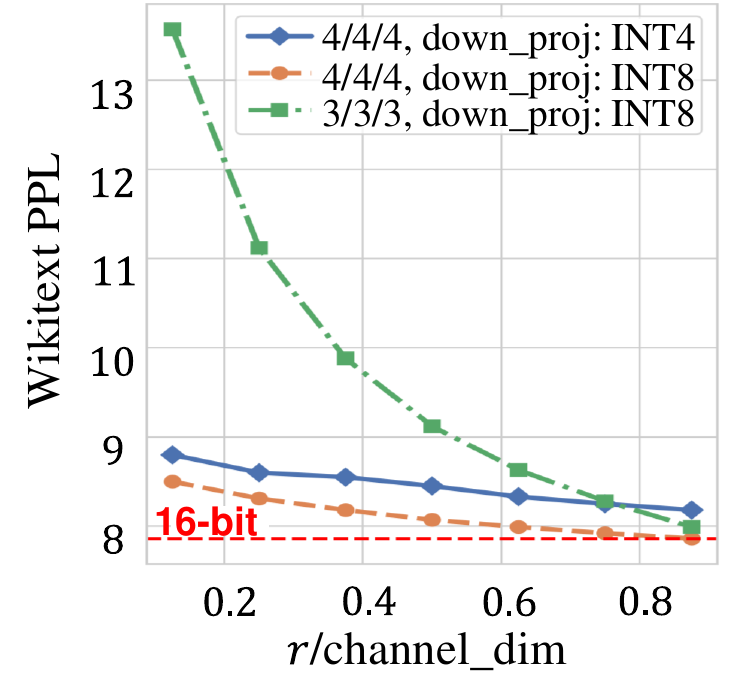}
		\caption{}
            \label{fig:ablation_rank}
	\end{subfigure}
	\centering
	\begin{subfigure}{0.235\textwidth}
		\includegraphics[width=\textwidth]{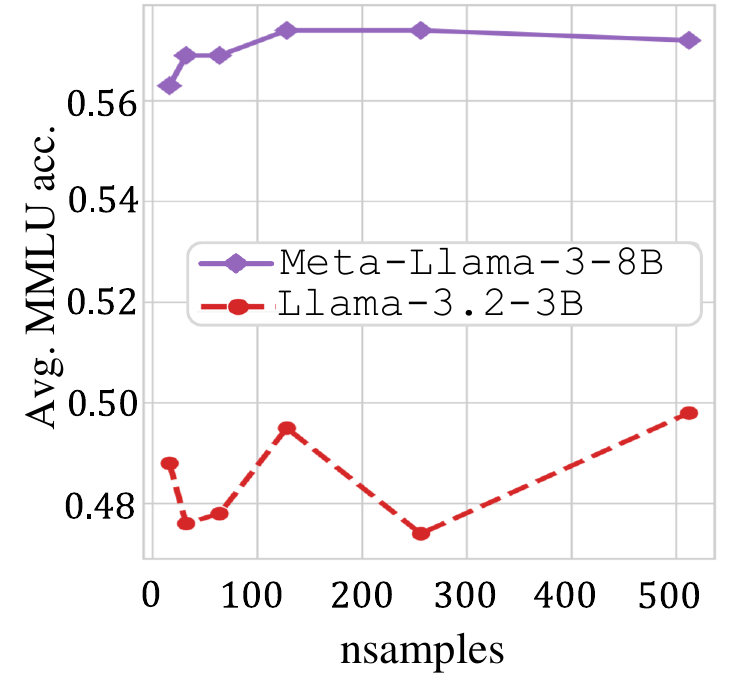}
		\caption{}
            \label{fig:ablation_nsamples}
	\end{subfigure} 
    \caption{Ablation study on (a) Changing rank of high precision subspace for \texttt{Llama-3.2-3B} and (b) Changing number of calibration samples.}
    \label{fig:quant_error}
\end{figure}
\subsection{Ablation studies}
\textbf{Projection bases}\ \ We evaluate the impact of different projections employed in ResQ by removing them and evaluating performance in Table \ref{tab:ablation_projections}. We see that removing ${\mU_D}$ or ${\mU_A}$ has a catastrophic impact on perplexity highlighting their importance. ${\mU_B}$ and ${\mU_C}$ which aid in quantization of KV cache have less severe impact when removed independently. But removing both of them leads to a non trivial increase in perplexity (particularly for \texttt{Meta-Llama-3-8B} and \texttt{Llama-3.2-3B} which employ grouped query attention).

\textbf{Rank of high-precision subspace}\ \ ResQ allows for seemless trade-off between accuracy and performance by modulating the rank $r$ of high precision subspace (Figure~\ref{fig:ablation_rank}). Increasing the rank improves perplexity albeit at the cost of increased computations in high precision. 

\textbf{Calibration dataset}\ \ We change number of Wikitext calibration samples used to obtain projections and evaluate performance in Figure~\ref{fig:ablation_nsamples}. For \texttt{Meta-Llama-3-8B}, MMLU accuracy increases with increasing samples and saturates beyond 128 samples. For \texttt{Llama-3.2-3B}, the trend is unclear with 512 samples achieving best performance. 

\section{Conclusion}
We introduce \emph{ResQ}, a novel mixed-precision, accelerator-friendly PTQ technique toward 4-bit quantization of large language models. ResQ projects weight, activation, and KV cache tensors to subspaces spanned by principal components, quantizing a low-rank ($\nicefrac{1}{8}$ of hidden dimension) high-variance subspace to 8-bit and the rest to 4-bit. ResQ outperforms both uniform- and mixed-precision quantization methods. We demonstrate the effectiveness of ResQ across a variety of tasks—including language modeling, language understanding, common-sense reasoning, language generation and multi modal understanding—using the Llama and Qwen models. Compared to SpinQuant, the strongest baseline, ResQ achieves up to 33\% lower perplexity on the WikiText dataset without requiring any additional training and offers up to $3.03\times$ speedup over the 16-bit baseline.

\section*{Impact Statement}
ResQ is a significant step forward towards efficiently serving LLMs in resource-constrained, on-device scenarios, potentially expanding the application space for these models. Although our approach aims to make LLMs more accessible and widely used, it does not address the potential risks of misuse for malicious purposes. To mitigate these risks, a strong commitment to user data protection, clear ethical guidelines, and transparency mechanisms is essential.
\section*{Acknowledgements}
The authors would like to thank Wanzin Yazar and Tristan Webb for infrastructure and technical assistance and Zifei Xu for helpful discussions. This work was supported by the Center for the Co-Design of Cognitive Systems (COCOSYS), a DARPA sponsored JUMP center of Semiconductor Research Corporation (SRC), Intel, SRC AIHW Program.
\nocite{langley00}
\balance
\bibliography{example_paper}
\bibliographystyle{icml2025}

\newpage
\appendix
\onecolumn

\section{Proof of Theorem \ref{theorem1}} \label{sec:proof_theorem}
We begin the proof by introducing the following lemma. 
\begin{lemma}
For any tensor $\mR$ quantized following the quantization described in equation \ref{eq:quant}, assuming the values of $\mR$ follows a normal distribution, we have 
\begin{equation}
    \mathbb{E}\lVert{\mR}-{Q(\mR)}\rVert_F \leq \frac{\sqrt{\pi\log\left[\text{size}(\mR)\right]}}{2^{n-1}-1}\mathbb{E}\lVert\mR\rVert_F
\end{equation}
where size($\mR$) denotes the number of elements in $\mR$. 
\label{lemma2}
\end{lemma}

Proof of lemma \ref{lemma2} can be found in \cite{li2024svdquantabsorbingoutlierslowrank}. From this lemma we obtain that the quantization error $\lVert\mR-Q(\mR)\rVert_F$ is bounded by the magnitude of the tensor quantized $\lVert\mR\rVert_F$. Now for our use case of mixed precision quantization where the low-precision component is quantized to $L$ bits and high precision component is quantized to $H$ bits, we write the quantization error again below, 
\begin{equation}
    \begin{split}
        \mathbb{E}\lVert{\mX} - {\mX_q}\rVert_F &= \mathbb{E}\lVert{\mX\mU_l} - Q_L({\mX\mU_l})\rVert_F \\
        &+ \mathbb{E}\lVert{\mX\mU_h} - Q_H({\mX\mU_h})\rVert_F .
    \end{split}
    \label{eq:quant_error_appendix}
\end{equation}
The random rotation matrices $\mR$ ensure that $\mX\mU_l$ and $\mX\mU_h$ are normally distributed by Lemma \ref{lemma1}. Applying Lemma \ref{lemma2} to the quantization error in equation \ref{eq:quant_error_appendix}, we get,
\begin{equation}
    \small
    \begin{split}
        \lVert{\mX} - {\mX_q}\rVert_F &\leq \frac{\sqrt{\text{log(size}(\mX\mU_l))\pi}}{2^{L-1}-1}\mathbb{E}||\mX\mU_l||_F \\
        &+ \frac{\sqrt{\text{log(size}(\mX\mU_h))\pi}}{2^{H-1}-1}\mathbb{E}||\mX\mU_h||_F \\
        &= \frac{\sqrt{\text{log(size}(\mX\mP_l))\pi}}{2^{L-1}-1}\mathbb{E}||\mX\mP_l||_F \\
        &+ \frac{\sqrt{\text{log(size}(\mX\mP_h))\pi}}{2^{H-1}-1}\mathbb{E}||\mX\mP_h||_F \\
        &= \frac{\sqrt{\text{log(size}(\mX\mP_l))\pi}}{2^{L-1}-1}\mathbb{E}||\text{tr}(\mX\mP_l\mP^\top_l\mX^\top)||_F \\
        &+ \frac{\sqrt{\text{log(size}(\mX\mP_h)\pi}}{2^{H-1}-1}\mathbb{E}||\text{tr}(\mX\mP_h\mP^\top_h\mX^\top)||_F 
    \end{split}
    \label{}
\end{equation}

We know $\text{size}(\mX\mP_l) = d-r$ and $\text{size}(\mX\mP_h) = r$ since $r$ components are in high precision. With $\mP_l\mP^\top_L + \mP_h\mP^\top_h = \mI$, we have
\begin{equation}
    \small
    \begin{split}
        \lVert{\mX}-{\mX_q}\rVert_F &\leq \frac{\sqrt{\text{log(d-r)}\pi}}{2^{L-1}-1}(\mathbb{E}\lVert{\mX}\rVert_F - \mathbb{E}\lVert{\mX\mP_h}\rVert_F) \\
        &+ \frac{\sqrt{\text{log(r)}\pi}}{2^{H-1}-1}\mathbb{E}\lVert{\mX\mP_h}\rVert_F \\
        &=\frac{\sqrt{\text{log(d-r)}\pi}}{2^{L-1}-1}\mathbb{E}\lVert{\mX}\rVert_F \\
        &-(\frac{\sqrt{\text{log(d-r)}\pi}}{2^{L-1}-1} - \frac{\sqrt{\text{log(r)}\pi}}{2^{H-1}-1})\mathbb{E}\lVert{\mX\mP_h}\rVert_F
    \end{split}
\end{equation}

Since $\frac{\sqrt{\text{log(d-r)}\pi}}{2^{L-1}-1} - \frac{\sqrt{\text{log(r)}\pi}}{2^{H-1}-1} > 0$ the quantization error is reduced by maximizing $\lVert{\mX\mP_h}\rVert_F$

\begin{figure}[!t]
	\centering
	\begin{subfigure}{.23\textwidth}
		\includegraphics[width=\textwidth]{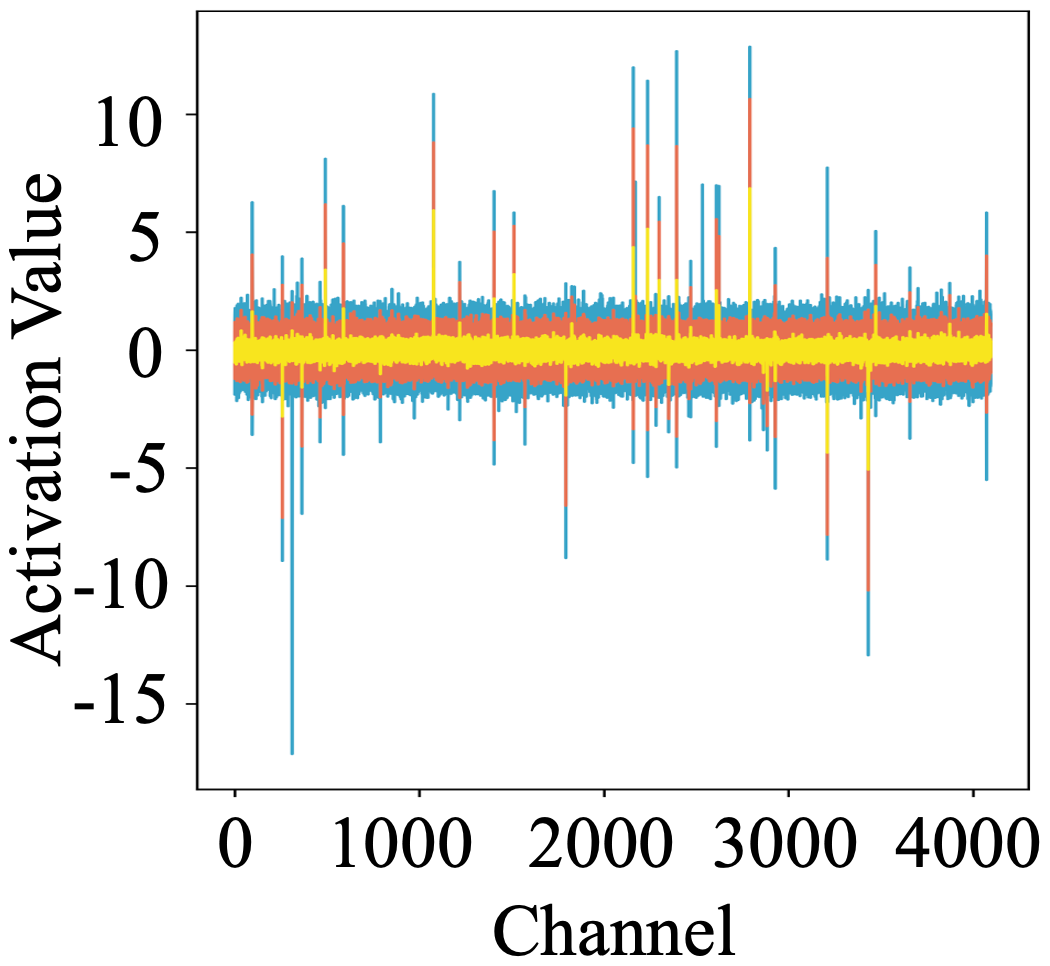}
		\caption{Baseline Attn Input}
	\end{subfigure}
	\centering
	\begin{subfigure}{.245\textwidth}
		\includegraphics[width=\textwidth]{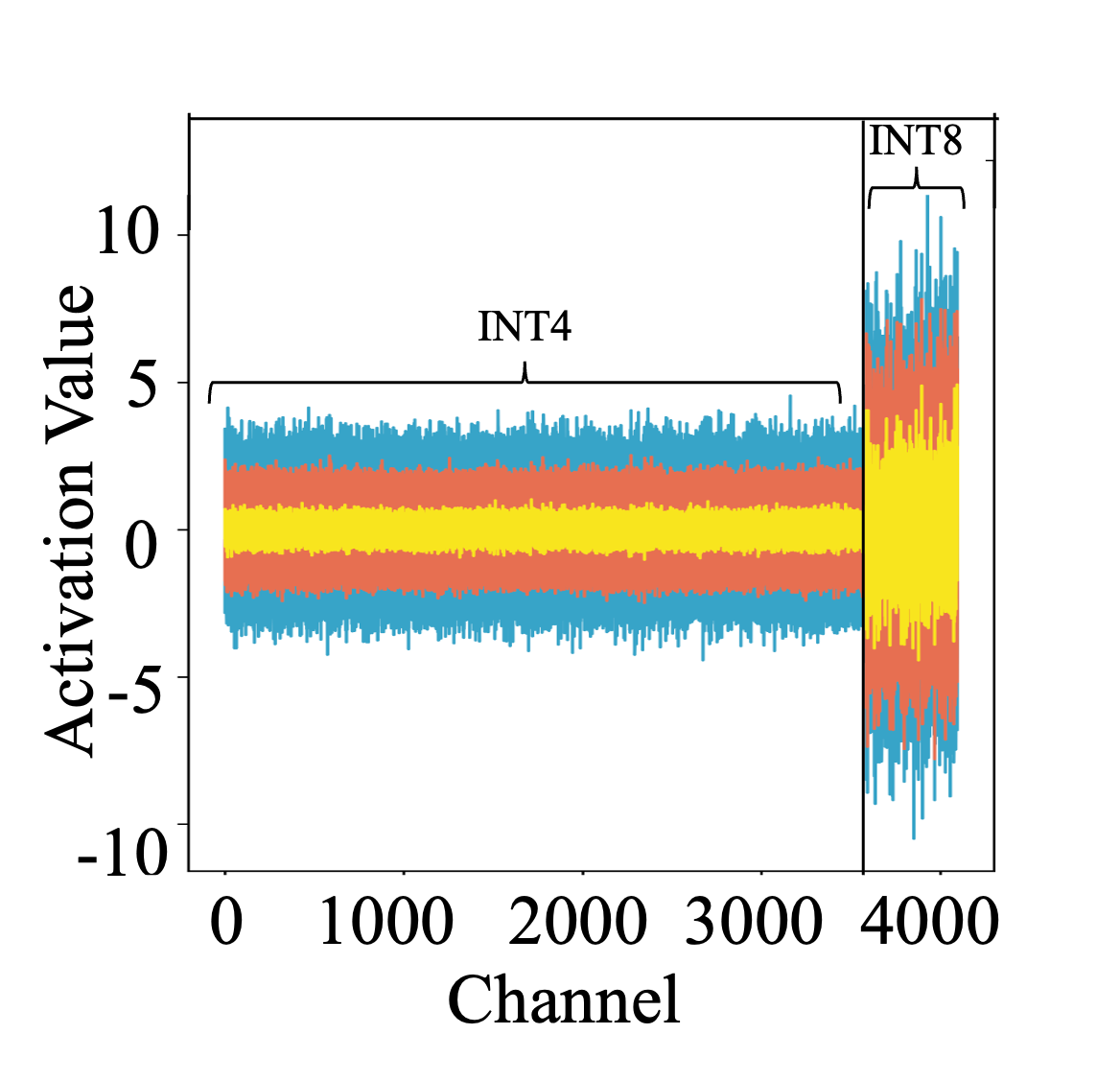}
		\caption{ResQ Attn Input}
	\end{subfigure} 
	\begin{subfigure}{0.24\textwidth}
            \includegraphics[width=\textwidth]{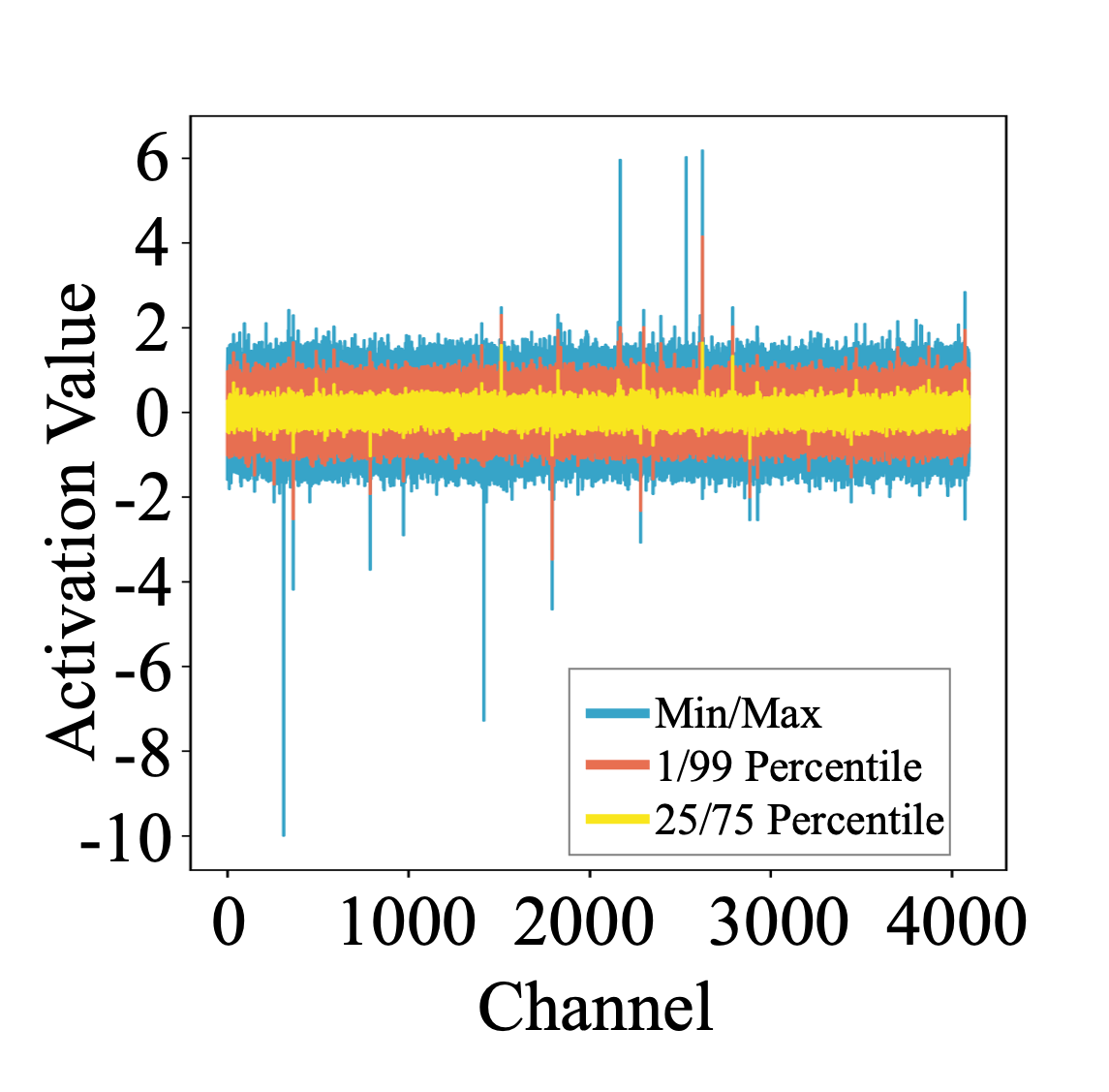}
    	\caption{Baseline FNN Input}
	\end{subfigure}
		\begin{subfigure}{0.235\textwidth}
            \includegraphics[width=\textwidth]{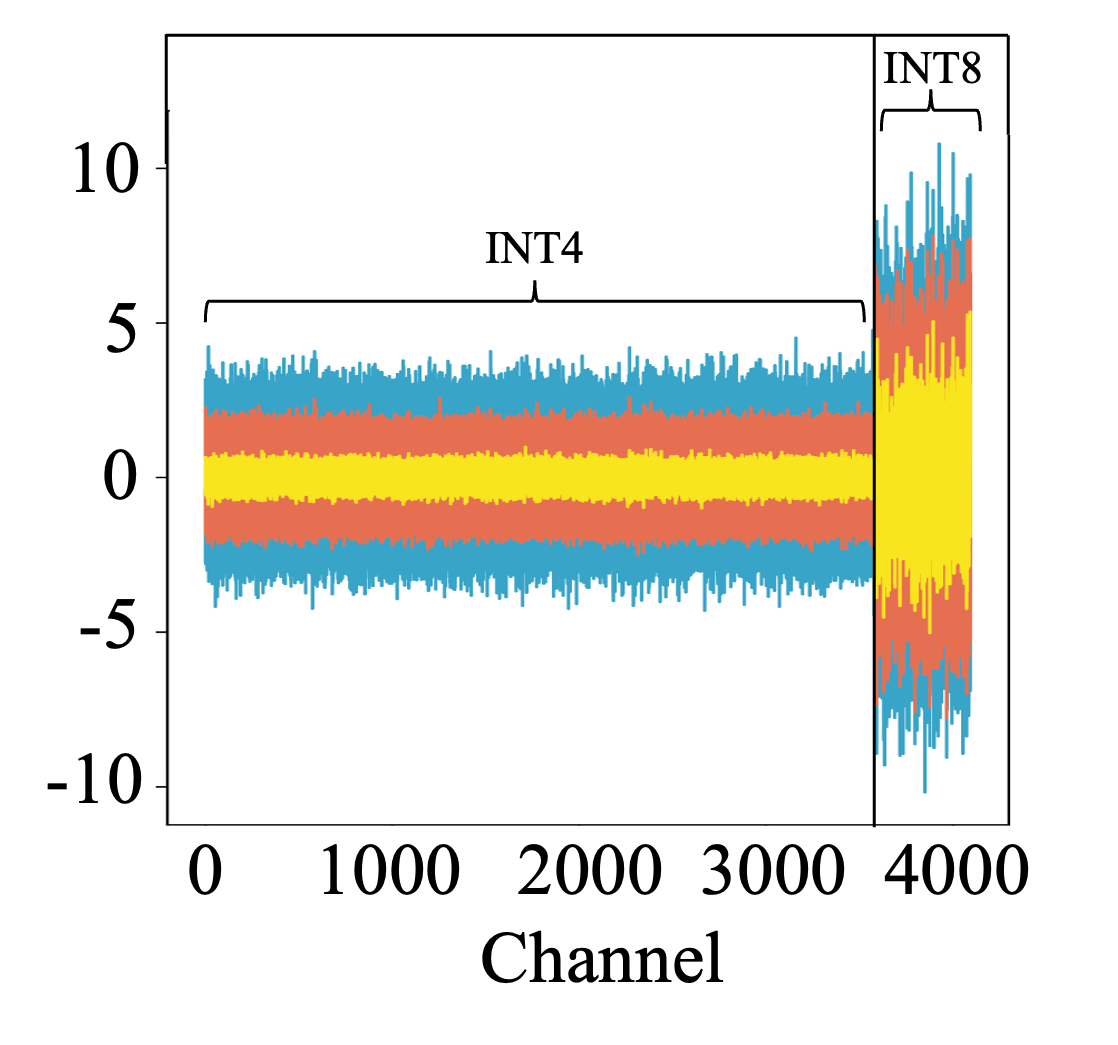}
    	\caption{ResQ FFN Input}
	\end{subfigure}
    \caption{Input activation distributions of attention and FFN layers, for baseline (a and c) and ResQ (b and d).}
    \label{fig:act_distribution}
\end{figure}

\section{Distribution of activations} \label{sec:appendix_act_dist}
The distribution of activations after projection by $\mU$ is shown in Figure \ref{fig:act_distribution}. The formulation of $\mU$ ensures that the final $r$ channels in the activation map comprise of coefficients along bases with maximum activation variance. Consequently, keep those channels in high precision minimizes quantization error. The remaining channels are more amenable to quantization due to the application of random rotations which suppress outlier values.

\begin{table}[h!]
\centering
\vspace*{-2mm}
\caption{Time taken (in NVIDIA A100 GPU hours) to quantize the model. All approaches use GPTQ for weight quantization. SpinQuant uses 4 GPUs to optimize rotation matrices.}
\vspace*{-3mm}
\label{tab:time_taken}
\resizebox{0.45\columnwidth}{!}{
\begin{tabular}{l|c|c|c}
\hline
             & QuaRot & ResQ & SpinQuant \\ \hline
\texttt{Llama-3.2-1B} & 4m     & 7m   & 13m       \\
\texttt{Llama-3.2-3B} & 8m     & 16m  & 38m    \\
\texttt{Meta-Llama-3-8B}   & 17m    & 35m  & 1h41m \\
\texttt{Llama-2-7b-hf}   & 15m    & 33m  & 1h37m  \\
\texttt{Llama-2-13b-hf}  & 23m    & 1h   & 3h42m   \\ \hline
\end{tabular}
}
\end{table}
\section{Additional implementation details}\label{sec:implementation_details}
In this work, obtaining the projection matrices and quantization of weights for \textbf{all the models} is performed on a single NVIDIA A100 80GB GPUs. Time taken by ResQ compared with other approaches is shown in Table \ref{tab:time_taken}. Evaluation on various benchmarks for all the models is also done on a single NVIDIA A100 GPU with the sole exception of \texttt{Meta-Llama-3-70b} which requires 4 GPUs for evaluation. We use \texttt{lm\_evaluation\_harness} version 0.4.5 \cite{lm-eval-harness} and {LongBench} \cite{bai2024longbench} for all the evaluation tasks. 
For Arc-c/e, Hellaswag, OpenBook QA, PIQA  tasks we report \texttt{acc\_norm} while for BoolQ, SIQA and Winogrande we report \texttt{acc}. 

For calibration data, we use 512 randomly choses samples for Wikitext to obtain the projection matrices. While for GPTQ we use 128 randomly choses samples from Wiktiext following the original work \cite{frantar2022gptq}. 

The KV cache, as well as the weights and activations of all Linear layers (except \texttt{mlp.down\_proj}), are quantized to 4-bit precision, with $\frac{1}{8}$ of channels retained in 8-bit precision. While, the weights and activations within \texttt{down\_proj} are uniformly quantized to 4-bit precision. Following \cite{ashkboos2024quarot} and \cite{liu2024spinquant}, we keep query vector in 16-bit.

\section{Complete results of main result tables} 
\label{sec:appendix_complete_results}
Detailed results of Table \ref{tab:main_results} in the main paper, including more models and task-by-task performance, are shown in Tables \ref{tab:llama2-7b_llama2-13b_eval_harness} (Llama families) and \ref{tab:Qwen-A4W4KV4} (Qwen2.5 family). As expected, ResQ achieves superior performance to baselines across the series of common sense reasoning and MMLU tasks.

\begin{table*}[ht!] 
\centering
\caption{Comparison of perplexity on Wikitext, accuracy on eight 0-shot common sense reasoning tasks including ARC-challenge, ARC-easy, BoolQ, HellaSwag, Openbook QA, PIQA, SIQA, and WinoGrande, and 0-shot massive multitask language understanding tasks across four subjects: STEM, Humanities, Social Sciences, and MMLU-other, for the Llama 2, Llama 3 and Llama 3.2 families when quantized to \textbf{W/A/KV = 4/4/4} bits. Results of all techniques were obtained using their official codebase. Our work ResQ and QUIK~\cite{ashkboos2023quik} keep $\nicefrac{1}{8}$ of channels in 8-bit. All techniques except RTN use GPTQ~\cite{frantar2022gptq}. ($\downarrow$): lower is better, ($\uparrow$): higher is better. }
\label{tab:llama2-7b_llama2-13b_eval_harness}
\resizebox{0.98\textwidth}{!}{
\begin{tabular}{c|c||c|ccccccccc|ccccc}
    \toprule
     \multicolumn{16}{c}{Llama 2 family}\\ \hline
     \multirow{3}{*}{Model}  & \multirow{3}{*}{Method} & Perplexity & \multicolumn{9}{c|}{0-shot common sense reasoning tasks} & \multicolumn{5}{c}{0-shot MMLU tasks} \\ \cline{3-17}
        &  & Wiki & ARC-c & ARC-e & BoolQ & HellaS & OBQA & PIQA & SIQA & WinoG & Avg. & humanities & Other & SocialS & STEM & Avg. \\
        &  & ($\downarrow$) & ($\uparrow$) & ($\uparrow$) & ($\uparrow$) & ($\uparrow$) & ($\uparrow$) & ($\uparrow$) & ($\uparrow$) & ($\uparrow$) & ($\uparrow$) & ($\uparrow$) & ($\uparrow$) & ($\uparrow$) & ($\uparrow$) & ($\uparrow$) \\ \hline 
\multirow{8}{*}{\texttt{Llama-2-7b-hf}} 
        & 16-bit      & 5.5 & 46.3 & 74.6 & 77.8 & 75.9 & 44.2 & 79.2 & 46.1 & 69.1 & 64.1 & 38.9 & 45.9 & 46.0 & 33.4 & 41.1 \\ \cdashline{2-17}
        & RTN         & 1766.2 & 26.3 & 27.8 & 54.8 & 29.4 & 25.8 & 51.0 & 35.0 & 48.7 & 37.4 & 24.5 & 24.7 & 22.9 & 22.2 & 23.6 \\
        & GPTQ        & 9600.0 & 24.8 & 31.4 & 55.4 & 30.6 & 25.6 & 55.8 & 34.2 & 53.3 & 38.9 & 24.7 & 24.5 & 22.7 & 23.2 & 23.8 \\
        & SmoothQuant+ & 15.4 & 29.3 & 47.1 & 56.8 & 48.6 & 31.8 & 65.5 & 37.2 & 52.4 & 46.1 & 25.0 & 24.5 & 24.1 & 23.4 & 24.2 \\
        & QUIK        & 7.5 & 39.8 & 63.7 & 68.9 & 68.3 & 37.8 & 72.9 & 42.1 & 62.4 & 57.0 & 26.9 & 29.6 & 28.8 & 25.8 & 27.8 \\
        & QuaRot      & 6.1 & 41.5 & 71.4 & 73.2 & 73.2 & 40.6 & 76.9 & 43.6 & 65.6 & 60.7 & 31.2 & 35.1 & 34.6 & 28.2 & 32.3 \\
        & SpinQuant   & 6.0 & 43.6 & 71.3 & 73.8 & 73.2 & 40.4 & 76.0 & \textbf{44.1} & 65.4 & 61.0 & 33.9 & 38.5 & 37.5 & 29.5 & 34.8 \\
        & \cellcolor[HTML]{CCFACC}ResQ        &\cellcolor[HTML]{CCFACC} \textbf{5.8}  & \cellcolor[HTML]{CCFACC}\textbf{44.0} & \cellcolor[HTML]{CCFACC}\textbf{72.6} & \cellcolor[HTML]{CCFACC}\textbf{75.3} & \cellcolor[HTML]{CCFACC}\textbf{74.0} & \cellcolor[HTML]{CCFACC}\textbf{41.0} & \cellcolor[HTML]{CCFACC}\textbf{77.9} & \cellcolor[HTML]{CCFACC}43.9 & \cellcolor[HTML]{CCFACC}\textbf{66.9} & \cellcolor[HTML]{CCFACC}\textbf{62.0} & \cellcolor[HTML]{CCFACC}\textbf{35.9} & \cellcolor[HTML]{CCFACC}\textbf{40.9} & \cellcolor[HTML]{CCFACC}\textbf{42.2} & \cellcolor[HTML]{CCFACC}\textbf{32.2} & \cellcolor[HTML]{CCFACC}\textbf{37.7} \\ \hline 

\multirow{8}{*}{\texttt{Llama-2-13b-hf}} 
        & 16-bit      & 4.9 & 49.1 & 77.4 & 80.5 & 79.4 & 45.2 & 80.7 & 47.2 & 72.1 & 66.5 & 47.9 & 59.3 & 61.0 & 42.4 & 52.7 \\ \cdashline{2-17}
        & RTN         & 3543.9 & 22.8 & 29.8 & 40.2 & 26.6 & 27.8 & 51.4 & 35.6 & 50.6 & 33.5 & 23.7 & 25.0 & 23.1 & 22.6 & 23.6 \\
        & GPTQ        & 3120.0 & 23.6 & 31.1 & 38.7 & 27.2 & 26.8 & 53.6 & 35.8 & 49.8 & 33.8 & 25.0 & 25.4 & 23.7 & 25.1 & 24.8  \\
        & SmoothQuant+ & 11.2 & 34.5 & 55.6 & 62.9 & 62.5 & 32.4 & 70.1 & 38.7 & 55.6 & 51.0 & 25.7 & 26.1 & 27.3 & 27.3 & 26.6 \\
        & QUIK        & 6.8 & 43.7 & 68.0 & 71.3 & 73.3 & 40.0 & 75.7 & 45.1 & 64.6 & 60.2 & 34.7 & 40.6 & 39.8 & 31.8 & 36.7 \\
        & QuaRot      & 5.4 & 46.9 & 74.9 & 76.6 & 75.8 & 42.6 & 79.1 & 45.5 & 69.0 & 63.8 & 43.8 & 53.6 & 54.0 & 39.4 & 47.7 \\
        & SpinQuant   & 5.2 & 49.0 & \textbf{76.3} & 78.2 & 77.1 & 42.8 & \textbf{79.3} & 46.3 & 69.5 & 64.8 & 43.5 & 53.1 & 55.4 & 39.1 & 47.8 \\
        & \cellcolor[HTML]{CCFACC}ResQ        & \cellcolor[HTML]{CCFACC} \textbf{5.1} & \cellcolor[HTML]{CCFACC}\textbf{49.1} & \cellcolor[HTML]{CCFACC}76.1 & \cellcolor[HTML]{CCFACC}\textbf{79.7} & \cellcolor[HTML]{CCFACC}\textbf{77.9} & \cellcolor[HTML]{CCFACC}\textbf{43.6} & \cellcolor[HTML]{CCFACC}79.1 & \cellcolor[HTML]{CCFACC}\textbf{46.6 }& \cellcolor[HTML]{CCFACC}\textbf{69.9} & \cellcolor[HTML]{CCFACC}\textbf{65.2} & \cellcolor[HTML]{CCFACC}\textbf{45.3} & \cellcolor[HTML]{CCFACC}\textbf{56.0} & \cellcolor[HTML]{CCFACC}\textbf{58.0} & \cellcolor[HTML]{CCFACC}\textbf{41.0} & \cellcolor[HTML]{CCFACC}\textbf{50.1} \\ \hline \hline 
        
        \multicolumn{16}{c}{Llama 3 family}\\ \hline
        \multirow{3}{*}{Model}  & \multirow{3}{*}{Method} & Perplexity & \multicolumn{9}{c|}{0-shot common sense reasoning tasks} & \multicolumn{5}{c}{0-shot MMLU tasks} \\ \cline{3-17}
        &  & Wiki & ARC-c & ARC-e & BoolQ & HellaS & OBQA & PIQA & SIQA & WinoG & Avg. & humanities & Other & SocialS & STEM & Avg. \\
        &  & ($\downarrow$) & ($\uparrow$) & ($\uparrow$) & ($\uparrow$) & ($\uparrow$) & ($\uparrow$) & ($\uparrow$) & ($\uparrow$) & ($\uparrow$) & ($\uparrow$) & ($\uparrow$) & ($\uparrow$) & ($\uparrow$) & ($\uparrow$) & ($\uparrow$) \\ \hline 
\multirow{8}{*}{\texttt{Meta-Llama-3-8B}} 
        & 16-bit      & 6.1 & 53.2 & 77.1 & 81.1 & 79.2 & 44.8 & 80.9 & 47.0 & 73.4 & 67.1 & 55.0 & 70.6 & 73.2 & 53.7 & 63.1 \\ \cdashline{2-17}
        & RTN         & 218.9 & 25.3 & 34.9 & 44.2 & 38.3 & 27.8 & 56.5 & 36.8 & 50.8 & 39.3 & 24.7 & 25.1 & 23.3 & 21.4 & 23.6 \\
        & GPTQ        & 166.3 & 24.7 & 37.7 & 44.3 & 36.8 & 27.0 & 57.6 & 36.4 & 53.8 & 39.8 & 24.7 & 23.9 & 22.8 & 21.8 & 23.3 \\
        & SmoothQuant+ & 78.2 & 27.5 & 42.0 & 50.7 & 44.9 & 28.8 & 59.0 & 35.9 & 50.9 & 42.5 & 25.4 & 25.5 & 24.5 & 23.4 & 24.7 \\
        & QUIK        & 14.2 & 33.6 & 56.4 & 60.5 & 61.5 & 33.2 & 68.7 & 39.9 & 59.0 & 51.6 & 30.0 & 34.0 & 34.8 & 32.1 & 32.7 \\
        & QuaRot      & 7.8 & 45.1 & 70.4 & 73.8 & 74.7 & 42.6 & 76.6 & 45.1 & 68.5 & 62.1 & 47.8 & 59.1 & 61.4 & 44.3 & 53.2 \\
        & SpinQuant   & 7.4 & 48.0 & \textbf{75.4} & \textbf{75.8} & 75.4 & \textbf{43.8} & 77.5 & 45.0 & 69.2 & 63.8 & 49.8 & 63.3 & 65.0 & 46.8 & 56.2 \\
        & \cellcolor[HTML]{CCFACC}ResQ        & \cellcolor[HTML]{CCFACC} \textbf{7.1} & \cellcolor[HTML]{CCFACC}\textbf{49.2} & \cellcolor[HTML]{CCFACC}75.0 & \cellcolor[HTML]{CCFACC}72.5 & \cellcolor[HTML]{CCFACC}\textbf{76.5} & \cellcolor[HTML]{CCFACC}43.0 & \cellcolor[HTML]{CCFACC}\textbf{78.3} & \cellcolor[HTML]{CCFACC}\textbf{45.8} & \cellcolor[HTML]{CCFACC}\textbf{71.0} & \cellcolor[HTML]{CCFACC}\textbf{63.9} & \cellcolor[HTML]{CCFACC}\textbf{50.6} & \cellcolor[HTML]{CCFACC}\textbf{64.4} & \cellcolor[HTML]{CCFACC}\textbf{65.8} & \cellcolor[HTML]{CCFACC}\textbf{48.1} & \cellcolor[HTML]{CCFACC}\textbf{57.2} \\ \hline
\multirow{8}{*}{\texttt{Meta-Llama-3-70B}}         
        & 16-bit       & 2.9 & 64.2 & 85.9 & 85.3 & 84.9 & 48.6 & 84.4 & 50.8 & 80.6 & 73.1 & 67.6 & 81.5 & 86.8 & 68.4 & 76.1 \\ \cdashline{2-17}
        & RTN          & 452.7 & 32.6 & 50.3 & 54.2 & 41.3 & 31.6 & 64.8 & 35.9 & 53.2 & 45.5 & 24.5 & 23.8 & 22.3 & 22.1 & 23.2 \\
        & GPTQ         & 11655.0 & 25.9 & 26.0 & 37.9 & 26.2 & 28.6 & 50.4 & 34.3 & 49.9 & 34.9 & 27.1 & 24.3 & 24.0 & 26.5 & 25.5 \\
        & SmoothQuant+ & - & - & - & - & - & - & - & - & - & - & - & - & - & - & - \\
        & QUIK         & 8.0 & 44.5 & 68.9 & 60.7 & 75.0 & 36.4 & 76.1 & 43.2 & 60.4 & 58.2  & 46.6 & 56.4 & 58.0 & 43.6 & 51.1  \\
        & QuaRot       & 5.7 & 53.7 & 74.5 & 81.6 & 81.1 & \textbf{46.6} & 81.0 & 46.8 & 75.2 & 67.6 & 55.7 & 72.5 & 75.8 & 57.3 & 65.3\\
        & SpinQuant    & 6.2 & 52.0 & 77.3 & 81.7 & 75.6 & 43.8 & 78.8 & 43.4 & 72.8 & 65.7 & 50.7 & 67.0 & 68.1 & 51.9 & 59.4 \\
        & \cellcolor[HTML]{CCFACC}ResQ       &\cellcolor[HTML]{CCFACC} \textbf{4.1}  & \cellcolor[HTML]{CCFACC}\textbf{61.4} & \cellcolor[HTML]{CCFACC}\textbf{84.3} & \cellcolor[HTML]{CCFACC}\textbf{83.9} & \cellcolor[HTML]{CCFACC}\textbf{83.5} & \cellcolor[HTML]{CCFACC}46.0 & \cellcolor[HTML]{CCFACC}\cellcolor[HTML]{CCFACC}\textbf{83.1} & \cellcolor[HTML]{CCFACC}\textbf{48.6} & \cellcolor[HTML]{CCFACC}\textbf{78.3} & \cellcolor[HTML]{CCFACC}\textbf{71.1} & \cellcolor[HTML]{CCFACC}\textbf{64.9} & \cellcolor[HTML]{CCFACC}\textbf{79.9} & \cellcolor[HTML]{CCFACC}\textbf{84.9} & \cellcolor[HTML]{CCFACC}\textbf{66.1} & \cellcolor[HTML]{CCFACC}\textbf{74.0} \\ \hline \hline
        
\multicolumn{16}{c}{Llama 3.2 family}\\ \hline
        \multirow{3}{*}{Model} & \multirow{3}{*}{Method} & Perplexity & \multicolumn{9}{c|}{0-shot common sense reasoning tasks} & \multicolumn{5}{c}{0-shot MMLU tasks} \\ \cline{3-17}
        &  & Wiki & ARC-c & ARC-e & BoolQ & HellaS & OBQA & PIQA & SIQA & WinoG & Avg. & humanities & Other & SocialS & STEM & Avg. \\
        &  & ($\downarrow$) & ($\uparrow$) & ($\uparrow$) & ($\uparrow$) & ($\uparrow$) & ($\uparrow$) & ($\uparrow$) & ($\uparrow$) & ($\uparrow$) & ($\uparrow$) & ($\uparrow$) & ($\uparrow$) & ($\uparrow$) & ($\uparrow$) & ($\uparrow$) \\ \hline 
\multirow{8}{*}{\texttt{Llama-3.2-1B}} 
        & 16-bit      & 9.8 & 36.5 & 60.6 & 63.4 & 63.6 & 37.4 & 74.5 & 42.8 & 60.1 & 54.9 & 34.8 & 41.1 & 39.9 & 32.0 & 36.9 \\ \cdashline{2-17}
        & RTN         & 329.1 & 22.4 & 29.9 & 53.4 & 31.4 & 29.4 & 54.8 & 34.9 & 48.5 & 38.1 & 24.8 & 25.2 & 22.4 & 22.7 & 23.8 \\
        & GPTQ        & 108.9 & 24.7 & 32.7 & 52.3 & 30.7 & 23.6 & 54.3 & 34.4 & 51.1 & 38.0 & 24.7 & 25.1 & 25.5 & 24.5 & 24.9 \\
        & SmoothQuant+ & 228.9 & 23.3 & 30.1 & 52.9 & 31.3 & 26.6 & 54.2 & 34.5 & 51.2 & 38.0 & 23.9 & 24.1 & 25.0 & 23.5 & 24.1 \\
        & QUIK        & 21.8 & 27.4 & 46.0 & 55.0 & 46.0 & 26.4 & 62.4 & 38.6 & 52.6 & 44.3 & 25.6 & 25.6 & 24.6 & 24.5 & 25.1 \\
        & QuaRot      & 14.3 & 30.0 & 51.4 & 59.1 & 54.0 & \textbf{34.2} & 66.7 & 39.6 & 57.1 & 49.0 & 25.4 & 26.9 & 25.4 & 24.4 & 25.5 \\
        & SpinQuant   & 13.6 & 32.3 & 51.8 & \textbf{59.3} & 55.4 & 30.4 & 67.7 & 38.6 & 54.7 & 48.8 & 25.4 & 27.6 & 24.2 & 25.3 & 25.6 \\
        & \cellcolor[HTML]{CCFACC}ResQ & \cellcolor[HTML]{CCFACC} \textbf{12.4} & \cellcolor[HTML]{CCFACC}\textbf{34.0} & \cellcolor[HTML]{CCFACC}\textbf{54.2} & \cellcolor[HTML]{CCFACC}57.0 & \cellcolor[HTML]{CCFACC}\textbf{57.3} & \cellcolor[HTML]{CCFACC}31.2 & \cellcolor[HTML]{CCFACC}\textbf{69.4} & \cellcolor[HTML]{CCFACC}\textbf{41.0} & \cellcolor[HTML]{CCFACC}\textbf{56.8} & \cellcolor[HTML]{CCFACC}\textbf{50.1} & \cellcolor[HTML]{CCFACC}\textbf{28.3} & \cellcolor[HTML]{CCFACC}\textbf{30.5} & \cellcolor[HTML]{CCFACC}\textbf{31.3} & \cellcolor[HTML]{CCFACC}\textbf{27.6} & \cellcolor[HTML]{CCFACC}\textbf{29.4} \\ \hline
\multirow{8}{*}{\texttt{Llama-3.2-3B}}         
        & 16-bit      & 7.8 & 46.2 & 71.7	& 73.1 & 73.7 & 43.4 & 77.4 & 47.2 & 69.1 & 62.7 & 48.9 & 62.9 & 62.3 & 45.2 & 54.8 \\ \cdashline{2-17}
        & RTN         & 268.8 & 23.5 & 35.4	& 46.2 & 35.6 & 28.2 & 56.3 & 33.6 & 50.6 & 38.7 & 25.1	& 25.6 & 27.0 & 24.9 & 25.7 \\
        & GPTQ        & 178.3 & 27.0 & 27.0	& 48.8 & 44.4 & 27.8 & 59.1 & 37.1 & 51.5 & 40.3 & 24.9 & 24.5 & 25.7 & 24.0 & 24.8     \\ 
        & SmoothQuant+ & 96.1 & 25.3 & 33.1 & 47.8 & 37.7 & 25.2 & 56.2 & 35.8 & 50.9 & 39.0 & 25.4 & 26.6 & 26.4 & 25.3 & 25.9 \\
        & QUIK        & 15.8 & 32.9 & 50.1	& 52.6 & 59.1 & 33.2 & 68.7 & 40.3 & 53.0 & 48.8 & 29.0	& 33.2 & 31.9 & 30.3 & 31.1 \\
        & QuaRot      & 10.1 & 38.6 & 59.0	& 65.9 & 66.5 & 35.8 & 74.4 & 43.1 & \textbf{65.2} & 56.1 & 38.5	& 47.3 & 46.7 & 35.3 & 42.0 \\
        & SpinQuant   & 9.2 & 38.9 & 64.8	& 68.0 & 69.1 & \textbf{39.4} & 74.9 & 45.1 & 62.9 & 57.9 & 37.0 & 49.4 & 50.5 & 39.9 & 44.2 \\
        & \cellcolor[HTML]{CCFACC}ResQ & \cellcolor[HTML]{CCFACC}\textbf{8.8} & \cellcolor[HTML]{CCFACC}\textbf{43.1} & \cellcolor[HTML]{CCFACC}\textbf{65.6} & \cellcolor[HTML]{CCFACC}\textbf{68.8} & \cellcolor[HTML]{CCFACC}\textbf{70.5} & \cellcolor[HTML]{CCFACC}38.4 & \cellcolor[HTML]{CCFACC}\textbf{75.1} & \cellcolor[HTML]{CCFACC}\textbf{45.6} & \cellcolor[HTML]{CCFACC}64.8 & \cellcolor[HTML]{CCFACC}\textbf{59.0} & \cellcolor[HTML]{CCFACC}\textbf{44.7} & \cellcolor[HTML]{CCFACC}\textbf{57.0} & \cellcolor[HTML]{CCFACC}\textbf{56.5} & \cellcolor[HTML]{CCFACC}\textbf{41.0} & \cellcolor[HTML]{CCFACC}\textbf{49.8} \\ 
    \bottomrule

\end{tabular}
}

\end{table*}

\begin{table*}[ht!] 
\centering
\caption{Comparison of perplexity score on Wikitext, accuracy on eight 0-shot common sense reasoning tasks including ARC-challenge, ARC-easy, BoolQ, HellaSwag, Openbook QA, PIQA, SIQA, and WinoGrande, and 0-shot massive multitask language understanding tasks across four subjects: STEM, Humanities, Social Sciences, and MMLU-other, for the Qwen2.5 family when quantized to \textbf{W/A/KV = 4/4/4} bits. Results of all techniques were obtained using their official codebase. Our work ResQ and QUIK~\cite{ashkboos2023quik} keep $\nicefrac{1}{8}$ of channels in 8-bit. All techniques except RTN use GPTQ~\cite{frantar2022gptq}. ($\downarrow$): lower is better, ($\uparrow$): higher is better, $\ast$: In cases where Hadamard matrix does not exist at the MLP dimension, random orthogonal rotation is used instead.}
\label{tab:Qwen-A4W4KV4}
\resizebox{0.98\textwidth}{!}{
\begin{tabular}{c|c||c|ccccccccc|ccccc}
    \toprule
    \multicolumn{16}{c}{Qwen2.5 family}\\ \hline
    \multirow{3}{*}{Model} & \multirow{3}{*}{Method} & Perplexity & \multicolumn{9}{c|}{0-shot common sense reasoning tasks} & \multicolumn{5}{c}{0-shot MMLU tasks} \\ \cline{3-17}
        &  & Wiki & ARC-c & ARC-e & BoolQ & HellaS & OBQA & PIQA & SIQA & WinoG & Avg. & humanities & Other & SocialS & STEM & Avg. \\
        &  & ($\downarrow$) & ($\uparrow$) & ($\uparrow$) & ($\uparrow$) & ($\uparrow$) & ($\uparrow$) & ($\uparrow$) & ($\uparrow$) & ($\uparrow$) & ($\uparrow$) & ($\uparrow$) & ($\uparrow$) & ($\uparrow$) & ($\uparrow$) & ($\uparrow$) \\ \hline 
\multirow{7}{*}{\texttt{Qwen2.5-0.5B}}
        & 16-bit       & 13.1 & 31.9 & 58.4 & 62.1 & 52.1 & 35.0 & 69.7 & 44.3 & 57.1 & 51.3 & 42.2 & 53.2 & 55.5 & 41.5 & 48.1 \\ \cdashline{2-17}
        & RTN          & 23204.3 & 26.2 & 27.0 & 39.3 & 26.0 & 24.0 & 50.7 & 34.5 & 51.5 & 34.9 & 24.8 & 24.0 & 22.8 & 24.3 & 23.9 \\
        & GPTQ         & 16302.3 & 23.7 & 26.9 & 39.0 & 26.5 & 26.4 & 50.2 & 33.4 & 49.6 & 34.5 & 24.1 & 24.8 & 23.5 & 23.0 & 23.9 \\
        & SmoothQuant+ & 10053.9 & 25.9 & 26.3 & 39.9 & 27.2 & 25.4 & 47.1 & 35.9 & 49.6 & 34.7 & 24.5 & 24.7 & 21.5 & 22.1 & 23.2 \\
        & QUIK         & 38.6 & 24.5 & 38.6 & 48.0 & 36.9 & 28.4 & 58.1 & \textbf{36.4} & \textbf{51.9} & 40.4 & \textbf{26.3} & 25.9 & 23.6 & 24.2 & 25.0 \\
        & QuaRot$\ast$ & 219.9 & 25.4 & 36.6 & 45.0 & 28.9 & \textbf{28.6} & 54.1 & 32.9 & 51.7 & 37.9 & 24.4 & 24.0 & 23.0 & 23.5 & 23.7 \\
        & \cellcolor[HTML]{CCFACC}ResQ & \cellcolor[HTML]{CCFACC}\textbf{29.6} & \cellcolor[HTML]{CCFACC}\textbf{27.1} & \cellcolor[HTML]{CCFACC}\textbf{44.2} & \cellcolor[HTML]{CCFACC}\textbf{53.2} & \cellcolor[HTML]{CCFACC}\textbf{38.8} & \cellcolor[HTML]{CCFACC}28.0 & \cellcolor[HTML]{CCFACC}\textbf{61.9} & \cellcolor[HTML]{CCFACC}34.4 & \cellcolor[HTML]{CCFACC}51.3 & \cellcolor[HTML]{CCFACC}\textbf{42.4} & \cellcolor[HTML]{CCFACC}26.1 & \cellcolor[HTML]{CCFACC}\textbf{27.5} & \cellcolor[HTML]{CCFACC}\textbf{25.3} & \cellcolor[HTML]{CCFACC}\textbf{26.0} & \cellcolor[HTML]{CCFACC}\textbf{26.2} \\ \hline 
\multirow{7}{*}{\texttt{Qwen2.5-1.5B}}
        & 16-bit       & 9.3 & 45.1 & 72.1 & 72.9 & 67.7 & 40.2 & 76.3 & 48.8 & 63.7 & 60.8 & 53.5 & 65.5 & 70.6 & 52.8 & 60.6 \\ \cdashline{2-17}
        & RTN          & 14518.9 & 23.1 & 27.2 & 43.9 & 26.8 & 25.6 & 51.3 & 33.4 & 52.5 & 35.5 & 23.8 & 24.5 & 23.8 & 22.7 & 23.7 \\
        & GPTQ         & 25769.7 & 23.9 & 26.9 & 43.9 & 26.1 & 27.6 & 49.7 & 32.1 & 51.5 & 35.2 & 24.6 & 24.7 & 23.7 & 23.8 & 24.2 \\
        & SmoothQuant+ & 31655.9 & 25.0 & 26.2 & 39.9 & 26.0 & 26.0 & 50.8 & 32.1 & 49.0 & 34.4 & 25.5 & 24.4 & 22.7 & 22.4 & 23.8 \\
        & QUIK         & 6613.5 & 21.8 & 31.9 & 40.9 & 27.9 & 27.4 & 52.8 & 35.2 & 48.6 & 35.8 & 24.6 & 24.0 & 21.9 & 21.7 & 23.1 \\
        & QuaRot       & 6599.9 & 23.6 & 37.3 & 46.2 & 28.6 & 27.0 & 56.3 & 35.2 & 52.4 & 38.3 & 24.5 & 24.3 & 23.0 & 22.4 & 23.5 \\
        & \cellcolor[HTML]{CCFACC}ResQ & \cellcolor[HTML]{CCFACC}\textbf{12.5} & \cellcolor[HTML]{CCFACC}\textbf{38.7} & \cellcolor[HTML]{CCFACC}\textbf{64.1} & \cellcolor[HTML]{CCFACC}\textbf{65.7} & \cellcolor[HTML]{CCFACC}\textbf{61.4} & \cellcolor[HTML]{CCFACC}\textbf{37.8} & \cellcolor[HTML]{CCFACC}\textbf{71.6} & \cellcolor[HTML]{CCFACC}\textbf{42.7} & \cellcolor[HTML]{CCFACC}\textbf{60.1} & \cellcolor[HTML]{CCFACC}\textbf{55.3} & \cellcolor[HTML]{CCFACC}\textbf{43.2} & \cellcolor[HTML]{CCFACC}\textbf{54.4} & \cellcolor[HTML]{CCFACC}\textbf{54.9} & \cellcolor[HTML]{CCFACC}\textbf{41.5} & \cellcolor[HTML]{CCFACC}\textbf{48.5} \\ \hline 
\multirow{7}{*}{\texttt{Qwen2.5-3B}}
        & 16-bit       & 8.0 & 47.4 & 73.0 & 77.5 & 73.6 & 42.0 & 78.7 & 49.9 & 68.4 & 63.8 & 56.6 & 71.0 & 76.3 & 60.6 & 66.1 \\ \cdashline{2-17}
        & RTN          & 39033.0 & 25.6 & 25.8 & 41.7 & 26.3 & 27.4 & 49.5 & 33.1 & 51.4 & 35.1 & 24.5 & 24.4 & 22.8 & 21.9 & 23.4 \\
        & GPTQ         & 9977.8 & 26.0 & 26.7 & 41.5 & 26.7 & 28.2 & 51.5 & 31.9 & 48.3 & 35.1 & 24.3 & 23.8 & 22.8 & 21.8 & 23.2 \\
        & SmoothQuant+ & 73306.7 & 25.4 & 24.5 & 41.0 & 26.4 & 29.8 & 48.4 & 32.4 & 50.4 & 34.8 & 25.6 & 24.7 & 23.1 & 22.4 & 23.9 \\
        & QUIK         & 15.5 & 36.1 & 55.4 & 61.4 & 57.2 & 36.2 & 67.1 & 40.8 & 55.3 & 51.2 & 36.4 & 42.8 & 42.4 & 36.1 & 39.4 \\
        & QuaRot       & 68.8 & 32.4 & 53.1 & 51.6 & 49.2 & 33.4 & 66.7 & 39.3 & 56.4 & 47.7 & 28.1 & 32.0 & 28.9 & 26.6 & 28.9 \\
        & \cellcolor[HTML]{CCFACC}ResQ & \cellcolor[HTML]{CCFACC}\textbf{9.0} & \cellcolor[HTML]{CCFACC}\textbf{45.3} & \cellcolor[HTML]{CCFACC}\textbf{70.5} & \cellcolor[HTML]{CCFACC}\textbf{72.7} & \cellcolor[HTML]{CCFACC}\textbf{70.2} & \cellcolor[HTML]{CCFACC}\textbf{42.4} & \cellcolor[HTML]{CCFACC}\textbf{76.8} & \cellcolor[HTML]{CCFACC}\textbf{46.7} & \cellcolor[HTML]{CCFACC}\textbf{64.4} & \cellcolor[HTML]{CCFACC}\textbf{61.1} & \cellcolor[HTML]{CCFACC}\textbf{53.1} & \cellcolor[HTML]{CCFACC}\textbf{66.5} & \cellcolor[HTML]{CCFACC}\textbf{70.5} & \cellcolor[HTML]{CCFACC}\textbf{54.8} & \cellcolor[HTML]{CCFACC}\textbf{61.2} \\ \hline
\multirow{7}{*}{\texttt{Qwen2.5-7B}}         
        & 16-bit       & 6.8 & 51.2 & 77.6 & 84.7 & 78.9 & 47.2 & 80.0 & 54.8 & 73.2 & 68.4 & 62.6 & 76.7 & 82.6 & 70.1 & 73.0 \\ \cdashline{2-17}
        & RTN          & 24382.1 & 24.5 & 26.3 & 37.8 & 26.0 & 29.0 & 51.0 & 34.1 & 50.1 & 34.9 & 24.9 & 24.3 & 23.4 & 24.9 & 24.4 \\
        & GPTQ         & 13593.7 & 25.2 & 25.6 & 37.8 & 26.3 & 28.2 & 52.4 & 34.4 & 48.9 & 34.8 & 24.4 & 24.3 & 22.8 & 22.6 & 23.5 \\
        & SmoothQuant+ & 19088.7 & 26.3 & 25.2 & 39.8 & 26.4 & 27.6 & 52.7 & 33.5 & 52.0 & 35.4 & 25.1 & 25.4 & 22.6 & 24.1 & 24.3 \\
        & QUIK         & 260.3 & 29.5 & 42.4 & 51.7 & 36.3 & 28.2 & 59.6 & 34.5 & 49.6 & 41.5 & 24.3 & 26.9 & 23.1 & 23.8 & 24.6 \\
        & QuaRot$\ast$ & 4035.9 & 25.9 & 41.0 & 39.1 & 29.1 & 27.6 & 57.9 & 35.7 & 50.6 & 38.4 & 24.8 & 24.4 & 24.4 & 22.7 & 24.1 \\
        & \cellcolor[HTML]{CCFACC}ResQ & \cellcolor[HTML]{CCFACC}\textbf{8.2} & \cellcolor[HTML]{CCFACC}\textbf{49.0} & \cellcolor[HTML]{CCFACC}\textbf{74.7} & \cellcolor[HTML]{CCFACC}\textbf{81.4} & \cellcolor[HTML]{CCFACC}\textbf{75.7} & \cellcolor[HTML]{CCFACC}\textbf{45.0} & \cellcolor[HTML]{CCFACC}\textbf{78.9} & \cellcolor[HTML]{CCFACC}\textbf{49.4} & \cellcolor[HTML]{CCFACC}\textbf{68.2} & \cellcolor[HTML]{CCFACC}\textbf{65.3} & \cellcolor[HTML]{CCFACC}\textbf{57.8} & \cellcolor[HTML]{CCFACC}\textbf{74.4} & \cellcolor[HTML]{CCFACC}\textbf{79.3} & \cellcolor[HTML]{CCFACC}\textbf{64.5} & \cellcolor[HTML]{CCFACC}\textbf{69.0}\\ \hline 
\multirow{7}{*}{\texttt{Qwen2.5-14B}}
        & 16-bit       & 5.3 & 58.8 & 79.4 & 85.4 & 82.9 & 45.4 & 81.9 & 55.3 & 75.8 & 70.6 & 69.9 & 81.9 & 86.2 & 76.5 & 78.6 \\ \cdashline{2-17}
        & RTN          & 2715 & 21.6 & 32.7 & 51.5 & 29.6 & 25.8 & 52.6 & 33.2 & 51.7 & 37.3 & 25.3 & 23.2 & 26.0 & 25.3 & 24.9 \\
        & GPTQ         & 5100.3 & 23.8 & 29.1 & 47.7 & 30.1 & 27.6 & 51.3 & 34.6 & 51.2 & 36.9 & 25.1 & 24.7 & 25.1 & 24.3 & 24.8 \\
        & SmoothQuant+ & 1375.7 & 27.0 & 26.3 & 38.0 & 26.8 & 29.2 & 51.6 & 32.4 & 49.3 & 35.1 & 25.9 & 24.5 & 22.2 & 22.2 & 23.7 \\
        & QUIK         & 10.5 & 45.0 & 67.1 & 64.7 & 68.9 & 37.6 & 74.8 & 43.9 & 59.3 & 57.6 & 48.9 & 61.1 & 64.7 & 51.5 & 56.6 \\
        & QuaRot       & 6.8 & 54.8 & 79.6 & 79.9 & 78.7 & 44.0 & 79.5 & 49.9 & \textbf{70.7} & 67.1 & 60.9 & 75.1 & 80.2 & 67.3 & 70.9 \\
        & \cellcolor[HTML]{CCFACC}ResQ & \cellcolor[HTML]{CCFACC}\textbf{6.2} & \cellcolor[HTML]{CCFACC}\textbf{57.6} & \cellcolor[HTML]{CCFACC}\textbf{82.1} & \cellcolor[HTML]{CCFACC}\textbf{84.9} & \cellcolor[HTML]{CCFACC}\textbf{81.1} & \cellcolor[HTML]{CCFACC}\textbf{44.8} & \cellcolor[HTML]{CCFACC}\textbf{80.5} & \cellcolor[HTML]{CCFACC}\textbf{51.7} & \cellcolor[HTML]{CCFACC}70.6 & \cellcolor[HTML]{CCFACC}\textbf{69.2} & \cellcolor[HTML]{CCFACC}\textbf{65.2} & \cellcolor[HTML]{CCFACC}\textbf{78.4} & \cellcolor[HTML]{CCFACC}\textbf{83.4} & \cellcolor[HTML]{CCFACC}\textbf{71.5} & \cellcolor[HTML]{CCFACC}\textbf{74.6} \\ \hline
\multirow{7}{*}{\texttt{Qwen2.5-32B}}         
        & 16-bit       & 5.0 & 55.7 & 78.0 & 87.4 & 84.1 & 44.4 & 82.3 & 56.4 & 75.2 & 70.4 & 73.1 & 83.6 & 89.6 & 81.2 & 81.9 \\ \cdashline{2-17}
        & RTN          & 1847.4 & 24.3 & 35.3 & 51.4 & 31.9 & 27.0 & 52.8 & 34.1 & 51.4 & 38.5 & 24.5 & 25.1 & 25.3 & 24.3 & 24.8 \\
        & GPTQ         & 3891.1 & 25.4 & 35.4 & 48.5 & 31.8 & 27.0 & 53.8 & 35.8 & 50.5 & 38.5 & 25.9 & 24.8 & 23.6 & 24.0 & 24.6 \\ 
        & SmoothQuant+ & - & - & - & - & - & - & - & - & - & - & - & - & - & - & - \\
        & QUIK         & 9.6 & 41.0 & 64.6 & 74.9 & 72.0 & 39.6 & 75.8 & 44.5 & 60.2 & 59.1 & 54.7 & 66.8 & 71.3 & 58.8 & 62.9 \\
        & QuaRot       & 6.1 & 54.5 & 76.1 & 85.1 & 81.5 & 44.2 & 80.1 & 51.3 & 70.4 & 67.9 & 68.5 & 80.0 & 86.0 & 76.0 & 77.6 \\
        & \cellcolor[HTML]{CCFACC}ResQ & \cellcolor[HTML]{CCFACC}\textbf{5.6} & \cellcolor[HTML]{CCFACC}\textbf{55.1} & \cellcolor[HTML]{CCFACC}\textbf{78.4} & \cellcolor[HTML]{CCFACC}\textbf{86.0} & \cellcolor[HTML]{CCFACC}\textbf{82.5} & \cellcolor[HTML]{CCFACC}\textbf{45.4} & \cellcolor[HTML]{CCFACC}\textbf{81.1} & \cellcolor[HTML]{CCFACC}\textbf{53.9} & \cellcolor[HTML]{CCFACC}\textbf{74.0} & \cellcolor[HTML]{CCFACC}\textbf{69.5} & \cellcolor[HTML]{CCFACC}\textbf{70.3} & \cellcolor[HTML]{CCFACC}\textbf{82.3} & \cellcolor[HTML]{CCFACC}\textbf{87.9} & \cellcolor[HTML]{CCFACC}\textbf{78.9} & \cellcolor[HTML]{CCFACC}\textbf{79.8} \\ \hline
\multirow{7}{*}{\texttt{Qwen2.5-72B}}        
        & 16-bit       & 3.9 & 62.6 & 83.2 & 89.2 & 86.0 & 46.6 & 83.6 & 58.4 & 77.7 & 73.4 & 77.2 & 86.9 & 90.6 & 82.4 & 84.3 \\ \cdashline{2-17}
        & RTN          & 45412.7 & 25.9 & 26.3 & 38.0 & 25.9 & 25.2 & 50.0 & 34.2 & 48.7 & 34.3 & 25.5 & 24.2 & 23.0 & 23.2 & 24.0 \\
        & GPTQ         & 37967.2 & 25.4 & 25.8 & 38.1 & 25.6 & 26.6 & 51.2 & 34.2 & 49.4 & 34.5 & 25.1 & 24.0 & 21.9 & 22.2 & 23.3 \\ 
        & SmoothQuant+ & - & - & - & - & - & - & - & - & - & - & - & - & - & - & - \\
        & QUIK         & 8.3 & 45.1 & 68.1 & 77.2 & 77.2 & 39.0 & 77.4 & 45.6 & 65.6 & 61.9 & 60.2 & 74.3 & 77.5 & 65.3 & 69.3\\
        & QuaRot       & 4.9 & 55.8 & \textbf{81.1} & 87.5 & 84.0 & 45.2 & 81.7 & 52.5 & 74.5 & 70.3 & 71.4 & 84.2 & 87.7 & 77.1 & 80.1 \\ 
        & \cellcolor[HTML]{CCFACC}ResQ & \cellcolor[HTML]{CCFACC}\textbf{4.6} & \cellcolor[HTML]{CCFACC}\textbf{58.4} & \cellcolor[HTML]{CCFACC}80.9 & \cellcolor[HTML]{CCFACC}\textbf{88.4} & \cellcolor[HTML]{CCFACC}\textbf{84.9} & \cellcolor[HTML]{CCFACC}\textbf{48.2} & \cellcolor[HTML]{CCFACC}\textbf{82.6} & \cellcolor[HTML]{CCFACC}\textbf{55.5} & \cellcolor[HTML]{CCFACC}\textbf{77.0} & \cellcolor[HTML]{CCFACC}\textbf{72.0} & \cellcolor[HTML]{CCFACC}\textbf{72.8} & \cellcolor[HTML]{CCFACC}\textbf{84.6} & \cellcolor[HTML]{CCFACC}\textbf{89.0} & \cellcolor[HTML]{CCFACC}\textbf{79.5} & \cellcolor[HTML]{CCFACC}\textbf{81.5} \\ 
    \bottomrule
\end{tabular}
}
\end{table*}

\section{Additional quantization results: W/A/KV = 4/8/4 bits of precision} 
\label{sec:appendix_w4a8kv4}
This section presents additional comparisons between baselines and ResQ for the Llama family when quantized to \textbf{W/A/KV = 4/8/4} bits of precision. Across various MMLU tasks and perplexity evaluations on WikiText, ResQ consistently outperforms all baselines. For 0-shot common sense reasoning tasks, except for \texttt{Meta-Llama-3-8B}, ResQ achieves the best average performance. In the case of \texttt{Meta-Llama-3-8B}, ResQ is the second-best method, with Quarot performing marginally better by less than 0.2\%.

\section{Complete results of the MMMU benchmark} 
\label{sec:appendix_MMMU}
This section presents task-by-task results for the MMMU benchmark across six subjects—Art \& Design, Business, Science, Health \& Medicine, Humanities \& Social Science, and Tech \& Engineering—for the Qwen2 VL family when quantized to \textbf{W/A/KV = 4/4/4} bits and \textbf{W/A/KV = 4/8/4} bits of precision. On average, ResQ consistently outperforms all baselines across different models. Notably, the advantage of ResQ becomes more pronounced with larger models. For instance, for \texttt{Qwen2-VL-7B-Instruct} at \textbf{W/A/KV = 4/8/4} bits of precision, ResQ achieves an average accuracy score of $48.8$, significantly outperforming the next-best method, QUIK, which scores $26.4$, representing an $\sim85\%$ relative improvement.

\begin{table*}[ht!] 
\centering
\caption{Accuracy on eight 0-shot common sense reasoning tasks including ARC-challenge, ARC-easy, BoolQ, HellaSwag, Openbook QA, PIQA, SIQA, and WinoGrande and 0-shot massive multitask language understanding tasks across four subjects: STEM, Humanities, Social Sciences, and MMLU-other, for the Llama 2, Llama 3 and Llama 3.2 families when quantized to\textbf{W/A/KV = 4/8/4} bits. Results of all techniques were obtained using their official codebase. Our work ResQ and QUIK~\cite{ashkboos2023quik} keep $\nicefrac{1}{8}$ of channels in 8-bit. All techniques except RTN use GPTQ~\cite{frantar2022gptq}. ($\downarrow$): lower is better, ($\uparrow$): higher is better.}
\label{tab:llama-4-8-4}
\resizebox{0.98\textwidth}{!}{
\begin{tabular}{c|c||c|ccccccccc|ccccc}
    \toprule
    \multicolumn{16}{c}{Llama 2 family}\\ \hline
    \multirow{3}{*}{Model} & \multirow{3}{*}{Method} & Perplexity & \multicolumn{9}{c|}{0-shot common sense reasoning tasks} & \multicolumn{5}{c}{0-shot MMLU tasks} \\ \cline{3-17}
        &  & Wiki & ARC-c & ARC-e & BoolQ & HellaS & OBQA & PIQA & SIQA & WinoG & Avg. & humanities & Other & SocialS & STEM & Avg. \\
        &  & ($\downarrow$) & ($\uparrow$) & ($\uparrow$) & ($\uparrow$) & ($\uparrow$) & ($\uparrow$) & ($\uparrow$) & ($\uparrow$) & ($\uparrow$) & ($\uparrow$) & ($\uparrow$) & ($\uparrow$) & ($\uparrow$) & ($\uparrow$) & ($\uparrow$) \\ \hline 
\multirow{8}{*}{\texttt{Llama-2-7b-hf}} 
        & 16-bit       & 5.5 & 46.3 & 74.6 & 77.8 & 75.9 & 44.2 & 79.2 & 46.1 & 69.1 & 64.1 & 38.9 & 45.9 & 46.0 & 33.4 & 41.1 \\ \cdashline{2-17}
        & RTN          & 7.2 & 41.5 & 65.9 & 71.9 & 71.8 & 39.4 & 76.5 & 42.7 & 65.7 & 59.4 & 27.6 & 28.4 & 31.6 & 29.0 & 29.2 \\
        & GPTQ         & 11.8 & 42.5 & 71.3 & 69.9 & 73.6 & 43.2 & 77.4 & 44.9 & 68.9 & 61.5 & 28.0 & 32.3 & 32.1 & 28.4 & 30.2 \\
        & SmoothQuant+ & 6.8 & 41.9 & 69.1 & 70.7 & 72.9 & 40.2 & 77.1 & 32.7 & 66.9 & 58.9 & 28.1 & 30.0 & 28.6 & 27.1 & 28.5 \\
        & QUIK         & 5.7 & 43.9 & 73.4 & \textbf{77.3} & 74.2 & \textbf{44.6} & 78.2 & 44.3 & \textbf{68.9} & 63.1 & 35.8 & 39.3 & 40.1 & 30.4 & 36.4 \\
        & QuaRot       & 5.7 & 43.6 & 73.6 & 75.4 & 74.8 & 42.6 & 77.6 & 45.1 & 67.9 & 62.6 & 36.3 & 43.3 & 41.5 & 31.3 & 38.1 \\
        & SpinQuant    & 5.7 & 43.5 & 73.3 & 75.4 & 74.8 & 42.6 & 77.5 & 45.0 & 68.4 & 62.6 & 37.0 & 42.0 & 43.4 & 31.8 & 38.5 \\
        & \cellcolor[HTML]{CCFACC}ResQ        & \cellcolor[HTML]{CCFACC}\textbf{5.6}  & \cellcolor[HTML]{CCFACC}\textbf{46.3} & \cellcolor[HTML]{CCFACC}\textbf{74.5} & \cellcolor[HTML]{CCFACC}77.1          & \cellcolor[HTML]{CCFACC}\textbf{75.0} & \cellcolor[HTML]{CCFACC}42.8          & \cellcolor[HTML]{CCFACC}\textbf{78.9} & \cellcolor[HTML]{CCFACC}\textbf{45.6} & \cellcolor[HTML]{CCFACC}68.8          & \cellcolor[HTML]{CCFACC}\textbf{63.6} & \cellcolor[HTML]{CCFACC}\textbf{39.1} & \cellcolor[HTML]{CCFACC}\textbf{45.9} & \cellcolor[HTML]{CCFACC}\textbf{47.9} & \cellcolor[HTML]{CCFACC}\textbf{34.9} & \cellcolor[HTML]{CCFACC}\textbf{42.0} \\ \hline 

\multirow{8}{*}{\texttt{Llama-2-13b-hf}} 
        & 16-bit       & 4.9 & 49.1 & 77.4 & 80.5 & 79.4 & 45.2 & 80.7 & 47.2 & 72.1 & 66.5 & 47.9 & 59.3 & 61.0 & 42.4 & 52.7 \\ \cdashline{2-17}
        & RTN          & 6.9 & 41.8 & 65.2 & 70.8 & 66.5 & 37.8 & 76.0 & 42.5 & 63.9 & 58.0 & 37.5 & 42.8 & 43.8 & 31.8 & 39.0 \\
        & GPTQ         & 6.2 & 46.2 & 73.2 & 76.0 & 73.4 & 43.2 & 78.2 & 44.4 & 69.6 & 63.0 & 35.9 & 42.1 & 39.0 & 30.8 & 36.9 \\
        & SmoothQuant+ & 5.6 & 45.0 & 71.5 & 76.8 & 73.4 & 44.6 & 76.7 & 31.9 & 67.6 & 60.9 & 32.8 & 42.7 & 40.6 & 32.4 & 37.1 \\
        & QUIK         & \textbf{5.0} & 47.5 & 76.4 & 78.9 & 78.4 & 42.8 & \textbf{80.4} & 46.8 & \textbf{72.5} & 65.5 & 46.1 & 57.0 & 58.2 & 40.0 & 50.3 \\
        & QuaRot       & \textbf{5.0} & 48.6 & 77.0 & 78.9 & 78.2 & 44.2 & 80.3 & 46.3 & 72.2 & 65.7 & 46.5 & 56.8 & 58.0 & 40.1 & 50.4 \\
        & SpinQuant    & \textbf{5.0} & 48.3 & 76.4 & 80.4 & 78.1 & 43.8 & 79.8 & 46.7 & 71.1 & 65.6 & 46.7 & 57.1 & 58.3 & 40.1 & 50.5 \\
        & \cellcolor[HTML]{CCFACC}ResQ        & \cellcolor[HTML]{CCFACC}\textbf{5.0} & \cellcolor[HTML]{CCFACC}\textbf{49.0} & \cellcolor[HTML]{CCFACC}\textbf{77.1} & \cellcolor[HTML]{CCFACC}\textbf{80.6} & \cellcolor[HTML]{CCFACC}\textbf{78.9} & \cellcolor[HTML]{CCFACC}\textbf{45.4} & \cellcolor[HTML]{CCFACC}79.9 & \cellcolor[HTML]{CCFACC}\textbf{47.2} & \cellcolor[HTML]{CCFACC}72.3 & \cellcolor[HTML]{CCFACC}\textbf{66.3} & \cellcolor[HTML]{CCFACC}\textbf{47.6} & \cellcolor[HTML]{CCFACC}\cellcolor[HTML]{CCFACC}\textbf{58.2} & \cellcolor[HTML]{CCFACC}\textbf{59.9} & \cellcolor[HTML]{CCFACC}\textbf{41.7} & \cellcolor[HTML]{CCFACC}\textbf{51.9} \\ \hline \hline 

\multicolumn{16}{c}{Llama 3 family} \\ \hline
        \multirow{3}{*}{Model} & \multirow{3}{*}{Method} & Perplexity & \multicolumn{9}{c|}{0-shot common sense reasoning tasks} & \multicolumn{5}{c}{0-shot MMLU tasks} \\ \cline{3-17}
        &  & Wiki & ARC-c & ARC-e & BoolQ & HellaS & OBQA & PIQA & SIQA & WinoG & Avg. & humanities & Other & SocialS & STEM & Avg. \\
        &  & ($\downarrow$) & ($\uparrow$) & ($\uparrow$) & ($\uparrow$) & ($\uparrow$) & ($\uparrow$) & ($\uparrow$) & ($\uparrow$) & ($\uparrow$) & ($\uparrow$) & ($\uparrow$) & ($\uparrow$) & ($\uparrow$) & ($\uparrow$) & ($\uparrow$) \\ \hline 
\multirow{8}{*}{\texttt{Meta-Llama-3-8B}} 
        & 16-bit       & 6.1 & 53.2 & 77.1 & 81.1 & 79.2 & 44.8 & 80.9 & 47.0 & 73.4 & 67.1 & 55.0 & 70.6 & 73.2 & 53.7 & 63.1 \\ \cdashline{2-17}
        & RTN          & 8.5 & 47.8 & 72.3 & 72.1 & 75.3 & 43.0 & 78.2 & 44.8 & 71.5 & 63.1 & 46.3 & 59.5 & 61.9 & 45.7 & 53.4 \\
        & GPTQ         & 7.5 & 47.1 & 71.1 & 72.2 & 72.7 & 42.6 & 78.2 & 45.7 & 72.8 & 62.8 & 40.4 & 60.3 & 61.6 & 46.2 & 52.1 \\
        & SmoothQuant+ & 8.3 & 44.8 & 71.2 & 75.4 & 73.6 & 40.0 & 79.0 & 43.6 & 68.1 & 62.0 & 42.1 & 53.6 & 54.1 & 38.5 & 47.1 \\
        & QUIK         & 6.7 & 50.0 & 75.7 & 80.1 & 77.4 & \textbf{45.8} & 80.0 & 45.1 & \textbf{74.8} & 66.1 & 52.1 & 65.7 & 68.1 & 49.8 & 58.9 \\
        & QuaRot       & 6.7 & 51.6 & 78.5 & 80.0 & 77.7 & 45.2 & 79.8 & 46.4 & 73.1 & \textbf{66.5} & 51.6 & 66.8 & 68.5 & 48.7 & 58.9 \\
        & SpinQuant    & 6.6 & 50.0 & 77.2 & \textbf{80.3} & 77.9 & 44.0 & \textbf{80.7} & \textbf{46.6} & 72.8 & 66.2 & 52.5 & 67.2 & 68.1 & 49.5 & 59.3 \\
        & \cellcolor[HTML]{CCFACC}ResQ        & \cellcolor[HTML]{CCFACC}\textbf{6.5} & \cellcolor[HTML]{CCFACC}\textbf{54.3} & \cellcolor[HTML]{CCFACC}\textbf{78.6} & \cellcolor[HTML]{CCFACC}77.2 & \cellcolor[HTML]{CCFACC}\textbf{78.4} & \cellcolor[HTML]{CCFACC}44.0 & \cellcolor[HTML]{CCFACC}79.2 & \cellcolor[HTML]{CCFACC}46.3 & \cellcolor[HTML]{CCFACC}73.2 & \cellcolor[HTML]{CCFACC}66.4 & \cellcolor[HTML]{CCFACC}\textbf{53.6} & \cellcolor[HTML]{CCFACC}\textbf{68.6} & \cellcolor[HTML]{CCFACC}\textbf{70.0} & \cellcolor[HTML]{CCFACC}\textbf{52.0} & \cellcolor[HTML]{CCFACC}\textbf{61.0} \\ \hline
\multirow{8}{*}{\texttt{Meta-Llama-3-70B}}         
        & 16-bit       & 2.9 & 64.2 & 85.9 & 85.3 & 84.9 & 48.6 & 84.4 & 50.8 & 80.6 & 73.1 & 67.6 & 81.5 & 86.8 & 68.4 & 76.1 \\ \cdashline{2-17}
        & RTN          & 16499 & 26.5 & 25.7 & 37.8 & 26.4 & 29.0 & 51.1 & 34.6 & 53.0 & 35.5 & 25.4 & 25.9 & 22.5 & 22.7 & 24.1 \\
        & GPTQ         & 8586.4 & 26.5 & 24.9 & 38.1 & 26.4 & 29.4 & 51.9 & 34.9 & 49.4 & 35.2 & 25.7 & 23.6 & 22.5 & 23.4 & 23.8 \\
        & SmoothQuant+ & - & - & - & - & - & - & - & - & - & - & - & - & - & - & - \\
        & QUIK        & 3.7 & 60.3 & 82.0 & 83.5 & 83.5 & 45.4 & 82.4 & 47.8 & 78.1 & 70.4 & 65.2 & 79.0 & 84.1 & 65.2 & 73.4  \\
        & QuaRot       & 3.6 & 60.0 & 84.3 & \textbf{84.9} & 83.9 & \textbf{49.2} & 83.9 & 49.4 & 78.8 & 71.8 & 64.3 & 80.0 & \textbf{85.8} & 66.7 & 74.2 \\
        & SpinQuant    & - & - & - & - & - & - & - & - & - & - & - & - & - & - & - \\
        & \cellcolor[HTML]{CCFACC}ResQ       & \cellcolor[HTML]{CCFACC}\textbf{3.3} & \cellcolor[HTML]{CCFACC}\textbf{63.0} & \cellcolor[HTML]{CCFACC}\textbf{84.7} & \cellcolor[HTML]{CCFACC}84.4 & \cellcolor[HTML]{CCFACC}\textbf{84.4} & \cellcolor[HTML]{CCFACC}48.2 & \cellcolor[HTML]{CCFACC}\textbf{84.2} & \cellcolor[HTML]{CCFACC}\textbf{50.1} & \cellcolor[HTML]{CCFACC}\textbf{80.8} & \cellcolor[HTML]{CCFACC}\textbf{72.5} &\cellcolor[HTML]{CCFACC} \textbf{68.2} & \cellcolor[HTML]{CCFACC}\textbf{86.0} &\cellcolor[HTML]{CCFACC}80.8 &\cellcolor[HTML]{CCFACC} \textbf{66.8} &\cellcolor[HTML]{CCFACC} \textbf{75.4} \\ \hline \hline

\multicolumn{16}{c}{Llama 3.2 family}\\ \hline
        \multirow{3}{*}{Model} & \multirow{3}{*}{Method}  & Perplexity & \multicolumn{9}{c|}{0-shot common sense reasoning tasks} & \multicolumn{5}{c}{0-shot MMLU tasks} \\ \cline{3-17}
        &  & Wiki & ARC-c & ARC-e & BoolQ & HellaS & OBQA & PIQA & SIQA & WinoG & Avg. & humanities & Other & SocialS & STEM & Avg. \\
        &  & ($\downarrow$) & ($\uparrow$) & ($\uparrow$) & ($\uparrow$) & ($\uparrow$) & ($\uparrow$) & ($\uparrow$) & ($\uparrow$) & ($\uparrow$) & ($\uparrow$) & ($\uparrow$) & ($\uparrow$) & ($\uparrow$) & ($\uparrow$) & ($\uparrow$) \\ \hline 
\multirow{8}{*}{\texttt{Llama-3.2-1B}} 
        & 16-bit       & 9.8 & 36.5 & 60.6 & 63.4 & 63.6 & 37.4 & 74.5 & 42.8 & 60.1 & 54.9 & 34.8 & 41.1 & 39.9 & 32.0 & 36.9 \\ \cdashline{2-17}
        & RTN          & 16.6 & 30.6 & 46.7 & 61.9 & 55.2 & 32.4 & 66.7 & 38.0 & 56.7 & 48.5 & 25.1 & 26.7 & 26.1 & 25.2 & 25.8 \\
        & GPTQ         & 15.3 & 32.8 & 50.9 & 61.6 & 54.8 & 31.6 & 67.4 & 39.1 & 55.8 & 49.2 & 24.1 & 26.2 & 23.9 & 24.3 & 24.6 \\
        & SmoothQuant+ & 20.6 & 30.0 & 47.7 & 50.2 & 50.8 & 31.2 & 66.3 & 37.5 & 54.1 & 46.0 & 24.9 & 26.6 & 25.5 & 23.9 & 25.2 \\
        & QUIK         & 11.6 & \textbf{35.0} & 57.9 & \textbf{62.3} & 59.4 & 35.4 & 71.7 & \textbf{41.9} & 56.9 & 52.6 & 28.2 & 31.5 & 29.6 & 27.2 & 29.1 \\
        & QuaRot       & 11.1 & 34.1 & \textbf{58.8} & 52.3 & 59.8 & \textbf{36.4} & \textbf{72.3} & 41.2 & 58.6 & 51.7 & 28.3 & 30.6 & 29.5 & 26.4 & 28.7 \\
        & SpinQuant    & 11.1 & 34.2 & 55.6 & 61.8 & 60.0 & 35.0 & 72.1 & 40.8 & 57.6 & 52.1 & 26.1 & 27.7 & 27.4 & 24.0 & 26.3 \\
        & \cellcolor[HTML]{CCFACC}ResQ & \cellcolor[HTML]{CCFACC}\textbf{10.4} & \cellcolor[HTML]{CCFACC}34.8 & \cellcolor[HTML]{CCFACC}58.0 & \cellcolor[HTML]{CCFACC}62.2 & \cellcolor[HTML]{CCFACC}\textbf{61.1} & \cellcolor[HTML]{CCFACC}34.4 & \cellcolor[HTML]{CCFACC}\textbf{72.3} & \cellcolor[HTML]{CCFACC}41.8 & \cellcolor[HTML]{CCFACC}\textbf{60.1} & \cellcolor[HTML]{CCFACC}\textbf{53.1} & \cellcolor[HTML]{CCFACC}\textbf{31.3} & \cellcolor[HTML]{CCFACC}\textbf{36.0} & \cellcolor[HTML]{CCFACC}\textbf{36.2} & \cellcolor[HTML]{CCFACC}\textbf{31.0} & \cellcolor[HTML]{CCFACC}\textbf{33.6} \\ \hline

\multirow{8}{*}{\texttt{Llama-3.2-3B}}         
        & 16-bit       & 7.8 & 46.2 & 71.7 & 73.1 & 73.7 & 43.4 & 77.4 & 47.2 & 69.1 & 62.7 & 48.9 & 62.9 & 62.3 & 45.2 & 54.8 \\ \cdashline{2-17}
        & RTN          & 17.8 & 36.6 & 51.3 & 55.3 & 64.3 & 36.6 & 73.5 & 42.1 & 62.4 & 52.8 & 38.5 & 46.7 & 46.6 & 35.0 & 41.7 \\
        & GPTQ         & 14.1 & 37.0 & 59.9 & 57.3 & 62.9 & 36.8 & 74.3 & 41.9 & 64.4 & 54.3 & 37.1 & 47.7 & 46.2 & 36.5 & 41.9 \\ 
        & SmoothQuant+ & 12.7 & 37.0 & 54.5 & 53.3 & 61.9 & 34.8 & 71.2 & 41.6 & 63.2 & 52.2 & 31.4 & 37.6 & 40.7 & 32.4 & 35.5 \\
        & QUIK         & 8.6 & 42.1 & 65.9 & 71.8 & 71.7 & 40.0 & 76.0 & 44.6 & 66.7 & 59.8 & 45.2 & 57.2 & 57.8 & 40.7 & 50.2 \\
        & QuaRot       & 8.4 & 43.4 & 68.9 & 69.5 & 71.2 & 40.6 & 76.8 & \textbf{46.0} & 67.2 & 60.5 & 45.0 & 56.1 & 56.0 & 40.0 & 49.3 \\
        & SpinQuant    & 8.4 & 43.5 & 67.8 & 70.6 & 71.9 & 41.6 & \textbf{76.9} & 44.9 & 68.5 & 60.7 & 46.1 & 56.7 & 57.1 & 39.4 & 49.8 \\
        & \cellcolor[HTML]{CCFACC}ResQ & \cellcolor[HTML]{CCFACC}\textbf{8.1} & \cellcolor[HTML]{CCFACC}\textbf{44.4} & \cellcolor[HTML]{CCFACC}\textbf{69.4} & \cellcolor[HTML]{CCFACC}\textbf{72.4} & \cellcolor[HTML]{CCFACC}\textbf{72.2} & \cellcolor[HTML]{CCFACC}\textbf{41.8} & \cellcolor[HTML]{CCFACC}76.3 & \cellcolor[HTML]{CCFACC}45.2 & \cellcolor[HTML]{CCFACC}\textbf{69.1} & \cellcolor[HTML]{CCFACC}\textbf{61.3} & \cellcolor[HTML]{CCFACC}\textbf{48.2} & \cellcolor[HTML]{CCFACC}\textbf{61.1} & \cellcolor[HTML]{CCFACC}\textbf{59.8} & \cellcolor[HTML]{CCFACC}\textbf{44.5} & \cellcolor[HTML]{CCFACC}\textbf{53.4} \\ 
    \bottomrule

\end{tabular}
}

\end{table*}

\begin{table*}[ht!] 
\centering
\caption{Accuracy (higher is better) on 0-shot massive multi-discipline multimodal understanding and reasoning tasks across six subjects: Art \& Design, Business, Science, Health \& Medicine, Humanities \& Social Science, and Tech \& Engineering for the Qwen2 VL Instruct family when quantized to \textbf{W/A/KV = 4/4/4} bits and \textbf{W/A/KV = 4/8/4} bits. Results of all techniques were obtained using their official codebase. Our work ResQ and QUIK~\cite{ashkboos2023quik} keep $\nicefrac{1}{8}$ of channels in 8-bit. All techniques except RTN use GPTQ~\cite{frantar2022gptq}.}
\label{tab:appendix_vision_language}
\resizebox{0.75\textwidth}{!}
{ 
\begin{tabular}{c|c||ccccccc}
    \toprule 
\multicolumn{9}{c}{\texttt{Qwen2-VL-2B-Instruct}} \\ \hline
\multirow{2}{*}{\textbf{W/A/KV} (bit)}& \multirow{2}{*}{Method} & \multicolumn{7}{c}{0-shot MMMU tasks}\\ \cline{3-9}
 &  & Art-Design & Business & Science & Health & Humanities & Tech & Avg. \\ \hline 
16/16/16 & Baseline    & 56.7 & 36.0 & 37.3 & 50.8 & 26.0 & 31.0 & 39.6 \\ \cdashline{1-9}
\multirow{5}{*}{4/4/4}          
         & RTN         & 28.3 & 18.7 & 26.0 & 26.7 & 21.3 & \textbf{29.1} & 25.0 \\
         & GPTQ        & 28.3 & \textbf{27.3} & 27.0 & 29.0 & \textbf{26.7} & 27.6 & 27.7 \\
         & QUIK        & 25.8 & 26.0 & 26.7 & 29.2 & 26.0 & 24.3 & 26.3 \\
         & QuaRot      & 24.2 & 23.3 & 20.7 & 26.7 & 26.0 & 22.9 & 24.0 \\
         & \cellcolor[HTML]{CCFACC}ResQ & \cellcolor[HTML]{CCFACC}\textbf{38.3} & \cellcolor[HTML]{CCFACC}21.3 & \cellcolor[HTML]{CCFACC}\textbf{28.7} & \cellcolor[HTML]{CCFACC}\textbf{45.0} & \cellcolor[HTML]{CCFACC}21.3 & \cellcolor[HTML]{CCFACC}23.3 & \cellcolor[HTML]{CCFACC}\textbf{29.7} \\ \hline 
\multirow{5}{*}{4/8/4} 
        & RTN         & 27.5 & 21.3 & 27.3 & 24.2 & 21.3 & 27.6 & 24.9 \\
        & GPTQ        & 24.2 & 23.3 & 24.0 & 18.3 & 21.3 & 29.5 & 23.4 \\
        & QUIK        & 33.3 & 28.7 & 32.0 & 32.5 & 26.0 & 18.1 & 28.4 \\
        & QuaRot      & 20.0 & 24.7 & 30.0 & 26.7 & 26.0 & \textbf{31.4} & 26.5 \\
        & \cellcolor[HTML]{CCFACC}ResQ & \cellcolor[HTML]{CCFACC}\textbf{37.5} & \cellcolor[HTML]{CCFACC}\textbf{32.0} & \cellcolor[HTML]{CCFACC}\textbf{32.7} & \cellcolor[HTML]{CCFACC}\textbf{47.5} & \cellcolor[HTML]{CCFACC}\textbf{26.7} & \cellcolor[HTML]{CCFACC}27.6 & \cellcolor[HTML]{CCFACC}\textbf{34.0} \\ \hline \hline
\multicolumn{9}{c}{\texttt{Qwen2-VL-7B-Instruct}} \\ \hline
\multirow{2}{*}{\textbf{W/A/KV} (bit)}& \multirow{2}{*}{Method} & \multicolumn{7}{c}{0-shot MMMU tasks}\\ \cline{3-9}
 &  & Art-Design & Business & Science & Health & Humanities & Tech & Avg. \\ \hline 
16/16/16 & Baseline     & 68.3 & 41.3 & 54.7 & 68.3 & 38.7 & 38.1 & 51.6 \\ \cdashline{1-9}
\multirow{5}{*}{4/4/4}          
         & RTN         & 24.2 & 28.0 & 29.3 & 22.5 & 29.3 & 27.1 & 26.7 \\
         & GPTQ        & 21.7 & 26.0 & 25.3 & 28.3 & 24.7 & 23.3 & 24.9 \\
         & QUIK        & 30.8 & 30.0 & 32.0 & 26.7 & 28.0 & 26.2 & 28.9 \\
         & QuaRot      & 21.7 & 21.3 & 28.7 & 25.0 & 20.7 & 29.5 & 24.5 \\
         & \cellcolor[HTML]{CCFACC}ResQ & \cellcolor[HTML]{CCFACC}\textbf{65.0} & \cellcolor[HTML]{CCFACC}\textbf{39.3} & \cellcolor[HTML]{CCFACC}\textbf{45.3} & \cellcolor[HTML]{CCFACC}\textbf{61.7} & \cellcolor[HTML]{CCFACC}\textbf{34.0} & \cellcolor[HTML]{CCFACC}\textbf{36.7} & \cellcolor[HTML]{CCFACC}\textbf{47.0} \\ \hline 
\multirow{5}{*}{4/8/4} 
        & RTN         & 23.3 & 28.7 & 27.3 & 25.0 & 22.7 & 24.3 & 25.2 \\
        & GPTQ        & 20.8 & 23.3 & 30.0 & 19.2 & 24.0 & 28.6 & 24.3 \\
        & QUIK        & 25.0 & 23.3 & 31.3 & 26.7 & 25.3 & 26.7 & 26.4 \\
        & QuaRot      & 20.8 & 26.0 & 30.0 & 19.2 & 24.7 & 26.2 & 24.5 \\
        & \cellcolor[HTML]{CCFACC}ResQ        & \cellcolor[HTML]{CCFACC}\textbf{67.5} & \cellcolor[HTML]{CCFACC}\textbf{39.3} & \cellcolor[HTML]{CCFACC}\textbf{51.3} & \cellcolor[HTML]{CCFACC}\textbf{64.2} & \cellcolor[HTML]{CCFACC}\textbf{36.7} & \cellcolor[HTML]{CCFACC}\textbf{33.8} & \cellcolor[HTML]{CCFACC}\textbf{48.8} \\ \bottomrule
\end{tabular}
}
\end{table*}

\section{Artifact licenses}
According to their licenses, all language models used in the paper fall under acceptable use case. The licenses for the models are linked for perusal: \href{https://huggingface.co/meta-llama/Llama-2-7b-hf/blob/main/LICENSE.txt}{\texttt{Llama-2-7b-hf}}, \href{https://huggingface.co/meta-llama/Llama-2-13b-hf/blob/main/LICENSE.txt}{\texttt{Llama-2-13b-hf}}, \href{https://huggingface.co/meta-llama/Meta-Llama-3-8B/blob/main/LICENSE}{\texttt{Meta-Llama-3-8B}}, \href{https://huggingface.co/meta-llama/Meta-Llama-3-70B/blob/main/LICENSE}{\texttt{Meta-Llama-3-70B}},
\href{https://huggingface.co/meta-llama/Llama-3.2-1B/blob/main/LICENSE.txt}{\texttt{Llama-3.2-1B}},
\href{https://huggingface.co/meta-llama/Llama-3.2-3B/blob/main/LICENSE.txt}{\texttt{Llama-3.2-3B}},
\href{https://huggingface.co/Qwen/Qwen2.5-0.5B/blob/main/LICENSE}{\texttt{Qwen2.5-0.5B}},
\href{https://huggingface.co/Qwen/Qwen2.5-1.5B/blob/main/LICENSE}{\texttt{Qwen2.5-1.5B}},
\href{https://huggingface.co/Qwen/Qwen2.5-3B/blob/main/LICENSE}{\texttt{Qwen2.5-3B}},
\href{https://huggingface.co/Qwen/Qwen2.5-7B/blob/main/LICENSE}{\texttt{Qwen2.5-7B}},
\href{https://huggingface.co/Qwen/Qwen2.5-14B/blob/main/LICENSE}{\texttt{Qwen2.5-14B}},
\href{https://huggingface.co/Qwen/Qwen2.5-32B/blob/main/LICENSE}{\texttt{Qwen2.5-32B}}, 
\href{https://huggingface.co/Qwen/Qwen2.5-72B/blob/main/LICENSE}{\texttt{Qwen2.5-72B}},
\href{https://huggingface.co/Qwen/Qwen2-VL-7B-Instruct/blob/main/LICENSE}{\texttt{Qwen2-VL-7B-Instruct}}, and 
\href{https://huggingface.co/Qwen/Qwen2-VL-2B-Instruct/blob/main/LICENSE}{\texttt{Qwen2-VL-2B-Instruct}}.


\end{document}